%% file: main_arxiv.tex
\documentclass{article}
\usepackage{arxiv}
\usepackage[round]{natbib}
\renewcommand{\bibname}{References}
\renewcommand{\bibsection}{\subsubsection*{\bibname}}
\usepackage{times}
\usepackage{hyperref}
\usepackage{xcolor}
\hypersetup{
    colorlinks,
    linkcolor={red!50!black},
    citecolor={blue!50!black},
    urlcolor={blue!80!black}
}

\usepackage{url}

\usepackage{algorithm}
\usepackage{algorithmic}
\usepackage{url}            
\usepackage{booktabs}       
\usepackage{amsfonts}       
\usepackage{nicefrac}       
\usepackage{microtype}      
\usepackage{xcolor}         

\usepackage{amsmath}
\usepackage{amssymb}
\usepackage{mathtools}
\usepackage{amsthm}

\theoremstyle{plain}

\usepackage{graphicx}
\usepackage{subfigure}
\usepackage{multirow}
\newcommand{\nonono}[1]{}
\usepackage{subfigure}
\usepackage{smile}
\usepackage{setspace}
\AtBeginDocument{%
  \addtolength\abovedisplayskip{0.1\baselineskip}%
  \addtolength\belowdisplayskip{0.2\baselineskip}%
  \addtolength\abovedisplayshortskip{0.2\baselineskip}%
  \addtolength\belowdisplayshortskip{0.2\baselineskip}%
}

\begin{document}

\title{Score Matching-based Pseudolikelihood Estimation of Neural
Marked Spatio-Temporal Point Process with Uncertainty Quantification}

\date{Oct 24, 2023}

\author{Zichong Li\thanks{Li, Q. Xu, Z. Xu, Mei and Zhao are affiliated with Georgia Tech. Zha is affiliated with The Chinese University of Hong Kong. Correspondence to \url{zli911@gatech.edu}, \url{tourzhao@gatech.edu} and  \url{zhahy@cuhk.edu.cn}.}, Qunzhi Xu, Zhenghao Xu, Yajun Mei, Tuo Zhao, Hongyuan Zha}

\renewcommand{\shorttitle}{}

\maketitle

\begin{abstract}
Spatio-temporal point processes (STPPs) are potent mathematical tools for modeling and predicting events with both temporal and spatial features. Despite their versatility, most existing methods for learning STPPs either assume a restricted form of the spatio-temporal distribution, or suffer from inaccurate approximations of the intractable integral in the likelihood training objective. These issues typically arise from the normalization term of the probability density function. Moreover, current techniques fail to provide uncertainty quantification for model predictions, such as confidence intervals for the predicted event's arrival time and confidence regions for the event's location, which is crucial given the considerable randomness of the data. To tackle these challenges, we introduce SMASH: a Score MAtching-based pSeudolikeliHood estimator for learning marked STPPs with uncertainty quantification. Specifically, our framework adopts a normalization-free objective by estimating the pseudolikelihood of marked STPPs through score-matching and offers uncertainty quantification for the predicted event time, location and mark by computing confidence regions over the generated samples. The superior performance of our proposed framework is demonstrated through extensive experiments in both event prediction and uncertainty quantification.
\end{abstract}

\section{Introduction}

Spatio-temporal point processes (STPPs) are stochastic processes that model event occurrences in time and space, where each event is associated with both temporal and spatial features. STPPs are widely used in various fields, including ecology \citep{STPP_review2016}, physiology \citep{STPP_brain2012}, and epidemiology\citep{STPP_covid2021}, to model complex event sequences such as earthquakes, brain activities, and disease outbreaks. Classical STPPs \cite{STPP_review2006,STPP_review2016} capture relatively simple spatio-temporal patterns through combining a temporal point process model, such as Poisson process \citep{kingman1992poisson} and self-excitation process \citep{hawkes1971spectra}, with a pre-specified spatial distribution estimator. With the advent of neural networks, flexible and expressive neural STPPs have been developed to model much more complicated event dynamics, see  \citet{DeepSTPP_2022,DNSTPP_2023,DSTPP_2023,ODESTPP_2021,SpecSTPP_2022}.

In neural STPPs, event distributions are typically characterized through intensity functions parametrized by the neural network, which model the influence of past events on present occurrences.
Current methods predominantly estimate parameters by maximizing the log-likelihood of event time and location. However, computing such a likelihood involves integrating the intensity function over time and space, which is usually intractable due to the intricate form of the neural network. Therefore, they resort to numerical approximation methods such as Monte Carlo integration \citep{STPP_review2016,SpecSTPP_2022, DNSTPP_2023}, which inevitably introduce approximation errors that compromise prediction accuracy.
Various works have sought solutions for this complication.
For instance, \citet{ODESTPP_2021} utilize ODE with the continuous-time normalizing flow (CNFs) to model continuous transformation of the distribution, but they need to integrate over the ODE trajectory by inefficient numerical ODE solver.
\citet{SDE_2019, DeepSTPP_2022} bypass the approximation of the integral by assuming restricted forms of the distribution such as Gaussian mixture models, but failing to capture complex spatio-temporal dynamics.
\citet{DSTPP_2023} propose a diffusion-based STPP model to avoid integrals and flexibly learn event distributions. However, they suffer from inefficient training and non-trivial hyperparameter configuration.

Another limitation of neural STPPs is the absence of uncertainty quantification. Given the huge randomness of event times and locations, the variance of data points typically far exceeds the mean, rendering point estimates unreliable and insufficient. 
Thus, it is crucial to quantify the uncertainty associated with event predictions, such as providing confidence intervals for predicted event times and confidence regions for locations. 
Moreover, when dealing with marked STPPs, where discrete marks are associated with each event, we aim to match the predicted posterior distribution to the ground truth data.
Since the learned model often exhibits overconfidence or underconfidence, uncertainty quantification for event mark prediction is also necessary to enable users to make more informed decisions.
\citet{smurf2023} propose SMURF-THP for marked temporal point process and provides uncertainty quantification, which will be further discussed in Section 3.

In this work, we introduce SMASH: a Score MAtching-based pSeudolikeliHood estimator for learning marked STPPs, to address the above issues. Specifically, SMASH bypasses the difficulty in integral calculation by adopting a score-matching objective \cite{score_matching2005} for the conditional likelihood of event time and location, which matches the derivative of the log-likelihood (known as score) to the derivative of the log-density of the underlying (unknown) distribution.
Furthermore, SMASH is essentially a score-based generative model, which facilitates generating samples from the predicted event distribution, thereby enabling the quantification of uncertainty in predictions by computing confidence regions over the generated samples.

In summary, we make three primary contributions:

$\bullet$ We propose SMASH, a consistent estimator for marked STPPs that leverages a normalization-free score-based pseudolikelihood objective to bypass the intractable integral computation involved in log-likelihood estimation. 
SMASH parametrizes the conditional score function of event location and the joint intensity of event time and mark, capturing intricate spatio-temporal dynamics with discrete marks.

$\bullet$ We provide uncertainty quantification for event predictions beyond unreliable point estimation by computing confidence regions, a feature unexplored and unevaluated by existing STPP methods.
SMASH supports flexible generation of event time, mark and location via score-based sampling algorithm, offering high-quality samples.

$\bullet$ We validate the effectiveness of SMASH using multiple real-world datasets. Our results demonstrate that SMASH offers significant improvements over other baselines in uncertainty quantification of model predictions.

The rest of the paper is organized as follows: Section 2 introduces the background; Section 3 presents our proposed method; Section 4 presents the experimental results; Section 5 discusses the related work; Section 6 draws a brief conclusion.

\section{Background}
\label{sec:2}

We briefly review marked spatio-temporal point process (MSTPP), neural STPP, score matching, Langevin dynamics and uncertainty quantification in this section.

\textbf{$\bullet$ Marked Spatio-Temporal Point Process}  is a stochastic process whose realization consists of an ordered sequence of discrete events $S=\{(t_i,k_i,\bm{x}_i)\}_{i=1}^L$ with length $L,$ where $t_i\in [0,T]$ is the time of occurrence, $k_i\in\{1,\cdots,M\}$ is the discrete event mark/type and $\bm{x}_i\in \mathbb{R}^d$ is the location of the event that has occurred at time $t_i.$ Denote the history events up to time $t$ as $\mathcal{H}_{t} = \{(t_j,k_j,\bm{x}_j):t_j<t\}$, the events’ distribution in MSTPPs is usually characterized via the \textit{conditional intensity function} $\lambda(t,k,\bm{x}\given \mathcal{H}_{t_i})$.
For simplicity, we omit conditional dependency on the history in the following discussions and employ a superscript $i$ to signify the condition on $\mathcal{H}_{t_i}$.
The conditional intensity is then defined as
\begin{align}
\lambda^i(t,k,\bm{x})\triangleq \lambda(t,k,\bm{x}\given \mathcal{H}_{t_i}) = \lim_{\Delta t, \Delta \bm{x}\downarrow 0} \frac{\mathbb{P}^i(t_i \in [t,t+\Delta t], k_i=k, \bm{x}_i \in B(\bm{x},\Delta \bm{x}))}{|B(\bm{x},\Delta \bm{x})|\Delta t},
\end{align}
where $B(\bm{x},\Delta x)$ denotes a ball centered at $\bm{x}$ and with radius $\Delta \bm{x}$. The conditional intensity function describes the instantaneous probability that an event of mark $k$ occurs at time $t$ and location $\bm{x}$ given the events' history $\mathcal{H}_{t_i}.$
The generalized conditional probability of the $i$-th event given history $\mathcal{H}_{t_i}$ can then be expressed as 
\begin{align} \label{joint_prob}
    p^i(t,k,\bm{x}) = \lambda^i(t,k,\bm{x}) e^{-\int_{t_{i-1}}^t \int_{\mathbb{R}^d} \sum_{l=1}^M \lambda^i(\tau, l, \bm{s}) d\tau d\bm{s}},
\end{align}
where $t\in [t_{i-1},T]$. The log-likelihood of the event sequence $S$ is:
\begin{align} \label{LL}
    \ell(S) = &\sum_{i=1}^L \log \lambda^i(t_i,k_i, \bm{x}_i) - \int_0^{T}\int_{\mathbb{R}^d}\sum_{l=1}^M \lambda^i(\tau,l,\bm{s}) d\tau d\bm{s}.
\end{align}
The second term in the above equation serves as a normalization term for the probability density.

\textbf{$\bullet$ Neural STPP} employs deep neural networks to parameterize the conditional intensity function $\lambda^i(t, k, \bm{x})$.
For instance, \citet{SpecSTPP_2022} and \citet{DNSTPP_2023} leverage Multi-Layer Perceptron (MLP) to construct a non-stationary kernel for modeling the intricate inter-dependency within the intensity.
\citet{DeepSTPP_2022} utilize the Transformer architecture to encode event history into a latent variable, subsequently deriving the intensity using Radial Basis Function (RBF) kernels.
\citet{ODESTPP_2021} decompose spatial distribution from time and models the spatial distribution by CNFs, while the time distribution is learned through an ordinary differential equation (ODE) latent process.
These methods estimate the model parameters by maximizing the log-likelihood presented in Eq.~\ref{LL}, where the evaluation of the intractable integral is either avoided by assuming restricted forms of event distribution or computed by numerical method.

\textbf{$\bullet$ Score Matching} \citep{score_matching2005} is a technique for estimating the parameters of unnormalized probability density models. It minimizes the expected squared difference between the model's log-density gradient (also known as the score) and the ground truth log-density gradient, preventing the calculation of the normalization integral. 
Denoising score matching \citep{denoise2011} enhances the scalability of score matching by adding a noise term to the original data points and matching the model's score to the noisy data's score.
\citet{gsm2019} and \citet{gsm2022} further expand score matching to more general classes of domains.
Notably, \cite{meng2022concrete} introduce concrete score matching for discrete data, broadening the application of score-based models in discrete domains.

\textbf{$\bullet$ Langevin Dynamics} is a widely-used sampling technique for score-based models, which generates samples from a target distribution using only its score function by simulating a continuous-time stochastic process.
\citet{mirror} introduce Mirror Langevin Dynamics (MLD) as a variant specifically designed for sampling from distributions with constrained domains. 

\textbf{$\bullet$ Uncertainty Quantification} is an essential aspect of data modeling, measuring the variance and reliability of model predictions \citep{UQ_book}.
In the context of continuous variables, such as event time and location in STPPs, uncertainty is quantified by providing confidence intervals or regions that are associated with certain confidence levels.
The ideal model should nicely match the confidence level with the actual coverage of the generated confidence interval/regions; the closer the match, the more adeptly the model captures the underlying distribution of the event occurrences.
On the other hand, for discrete variables such as event marks, we quantify uncertainty by measuring the difference between the learned posterior distribution and the actual data distribution.
This involves a comparison between the predicted probability and the actual accuracy of the model \citep{calibration_ece}.

\section{Method}
\label{sec:3}
\subsection{Score Matching-based Pseudolikelihood}
Retaining the notation introduced in Section \ref{sec:2}, our goal is to learn $\lambda^i(t,k,\bm{x}),$ the joint intensity given the history $\mathcal{H}_{t_i}$.
We aim to employ the score matching technique to avoid the computation of the intractable integral in the log-likelihood (Eq. \ref{LL}).
However, score matching cannot be directly applied to the joint distribution of time, mark and location.
This is because: (1) an intractable integral on location space remains and (2) event mark is discrete and the gradient does not exist.
To tackle these issues, we decompose the joint intensity by conditioning on the event time and mark:
\begin{align}
\lambda^i(t,k,\bm{x}) = \lambda^i(t,k)p^i(\bm{x}\given t,k).
\end{align}
The first term is essentially the intensity function of a marked temporal point process (MTPP) without spatial features and the second term captures the spatial dynamics.
We parametrize both the marginal intensity and the spatial distribution, and derive two score matching-based objectives for estimation, which will be introduced in Section 3.1.1 and 3.1.2, respectively.

\subsubsection{Event time and mark} \label{sec:3.1.1}
We parametrize the marginal intensity function on event time and mark as $\lambda^i(t,k;\theta)$ using a Transformer model, where $\theta$ represents the model parameters.
For notational convenience, we omit $\theta$ in the subsequent context.
Since score matching cannot be directly applied to $p^i(t,k)$ due to the discrete nature of mark, we apply score matching to the conditional distribution of event time $p^i(t\given k)$ while adopting the conditional likelihood for event mark distribution $p^i(k\given t)$.
Here, the term "conditional" refers to the condition on time/mark besides the history.

The score function is defined as the gradient of log-probability density with respect to the data point.
As event time lies in the continuous domain, we have the score function of the conditional likelihood of event time given the mark $k$ and the history $\mathcal{H}_{t_i}$ as:
 \begin{align} \label{score_f}
\psi_t^i(t\given k)&=\partial_t \log p^i(t \given k)=\partial_t 
\log \lambda^i(t,k)-\sum_{l=1}^M \lambda^i(t,l).
\end{align}
We can then formulate the expected score matching objective for the $i$-th event time:
\begin{align} \label{eq:SM_one_event}
\mathcal{L}_{\rm time}^{(i)} = \frac{1}{2} \mathbb{E}_{t,k,\bm{x}}\left [\lVert\psi_t^i(t\given k)-{\psi_t^i}^*(t\given k)\rVert^2 \right],
\end{align}
where ${\psi_t^i}^*$ is the conditional score function of the true distribution of event time. However, this objective is unattainable as it depends on the unknown ground truth score. We can resolve this issue by following the general derivation in \cite{score_matching2005} and derive an empirical objective for the event sequence $S$:
\begin{align} \label{eq:SM}
    \widetilde{\mathcal{L}}_{\mathrm{time}}(S)=\sum_{i=1}^L \left[\frac{1}{2} {\psi_t^i(t_i \given k_i)}^2+\partial_t \psi_t^i(t_i \given k_i)\right].
\end{align}

By minimizing the above score matching objective, the scores of event time will be matched with the ground truth.
For the discrete event mark with an undefined score, we can derive the conditional probability mass function of the mark, given arrival time and history, in a closed form.
\begin{align}
p^i(k\given t) = \frac{\lambda^i(t,k)}{\sum_{l=1}^M \lambda^i(t,l)}.
\end{align}
Then we learn the event mark distribution by maximizing the conditional likelihood of $k$. This is equivalent to minimizing the following cross-entropy loss for the sequence $S$: 
\begin{align} \label{eq:CE}
\widetilde{\mathcal{L}}_{\rm mark}(S) = \sum_{i=1}^L - \log \frac{\lambda^i(t_i,k_i)}{\sum_{l=1}^M \lambda^i(t_i,l)}.
\end{align}

\subsubsection{Event location} \label{sec:3.1.2}
For the spatial distribution $p^i(\bm{x}|t,k)$, we parametrize the conditional score function $\psi_{\bm{x}}^i(\bm{x}\given t,k) = \nabla_{\bm{x}} \log p^i(\bm{x}|t,k)$.
This allows us to derive the empirical score matching objective for event location as:
\begin{align} \label{eq:SM_space}
    \widetilde{\mathcal{L}}_{\mathrm{spatial}}(S)=\frac{1}{L}\sum_{i=1}^L \biggl [\frac{1}{2} {\lVert\psi_{\bm{x}}^i(\bm{x}_i\given t_i, k_i)\rVert}^2 +  \sum_{j=1}^d \partial_{x^{(j)}} {\psi_{\bm{x}}^i}^{(j)}(\bm{x}_i \given t_i, k_i)\biggr ],
\end{align}
where $x^{(j)}$ is the j-th dimension of location and ${\psi_{\bm{x}}^i}^{(j)}$ is the j-th dimension of $\psi_{\bm{x}}^i$.

\subsubsection{Maximize Pseudolikelihood for Estimation}
Given the score matching-based objective for spatio-temporal features and the conditional log-likelihood for event mark, we can estimate the model by minimizing the pseudolikelihood represented as the weighted sum:
\begin{align}\label{eq:smash}
\widetilde{\mathcal{L}}_{\rm SMASH}(S) = \widetilde{\mathcal{L}}_{\rm time}(S)+\widetilde{\mathcal{L}}_{\rm spatial}(S) + \alpha  \widetilde{\mathcal{L}}_{\rm mark}(S),
\end{align}
where $\alpha$ is a hyperparameter that regulates the scale of the conditional log-likelihood objective.

\subsection{Denoising Score Matching}
In practice, the direct score-based objective given by Eq.~\ref{eq:smash} has two limitations: 1) the score functions of event location are inaccurately estimated in regions of low data density \citep{SMLD} and 2) the objective $\widetilde{\mathcal{L}}_{\rm time}(S)$ contains second-order derivatives, leading to numerical instability.
We resolve these issues by applying denoising score matching \cite{denoise2011} to the conditional distribution of event time and location, which reduces the area of the low-density region by perturbing data and improves stability by simplifying the objective.
Specifically, we add Gaussian noise to events' time and location.
Then we match the model to the perturbed data distribution $\tilde{p}^i(\tilde{t},k,\tilde{\bm{x}})=\int_t \int_{\bm{x}} {p}^i(t,k, \bm{x}) q(\tilde{t}\given t)q(\tilde{\bm{x}}\given \bm{x})dt d\bm{x}$, where $q(\tilde{t}\given t)\sim \mathcal{N}(t,\sigma_1)$ and $q(\tilde{\bm{x}}\given \bm{x})\sim \mathcal{N}(\bm{x},\bm{\sigma}_2)$.
Denote the perturbed $i$-th event as $\{(\tilde{t}_i^j,k,\tilde{\bm{x}}_i^j)\}_{j=1}^Q$, where $Q$ is the number of perturbed samples. The denoising variant of the score matching objective for event time and location can  be expressed as
\begin{align}
\widetilde{\mathcal{L}}_{\mathrm{time}}^{\rm D}(S)& = \frac{1}{2}\sum_{i,j=1}^{L,Q} \big [ \psi_{\tilde{t}}^i(\tilde{t}_i^j \given k)  -  \partial_{\tilde{t}} \log q(\tilde{t}_i^j \given  t_i)\big ]^2,\\
\widetilde{\mathcal{L}}_{\mathrm{spatial}}^{\rm D}(S)& = \frac{1}{2}\sum_{i,j=1}^{L,Q} \big [ \psi_{\tilde{\bm{x}}}^i(\tilde{\bm{x}}_i^j \given k)  -  \partial_{\tilde{\bm{x}}} \log q(\tilde{\bm{x}}_i^j \given  t_i)\big ]^2.
\end{align}
We then train the model by minimizing the following objective:
\begin{align} \label{DJSM:3}
    \widetilde{\mathcal{L}}_{\mathrm{SMASH}}^{\rm D}(S) =  \widetilde{\mathcal{L}}_{\mathrm{time}}^{\rm D}(S) + \widetilde{\mathcal{L}}_{\mathrm{spatial}}^{\rm D}(S) + \alpha \widetilde{\mathcal{L}}_{\mathrm{mark}}(S).
\end{align}

\subsection{Sampling}
With the learned intensities and score functions, we can generate new samples using Langevin Dynamics (LD) for event prediction and uncertainty quantification.
Suppose we want to generate the $i$-th event conditioning on the history $\mathcal{H}_{t_i}$, we first generate events' time and mark by sampling the initial time gap $t^{(0)}$ and event mark $k^{(0)}$ from a pre-specified distribution $\pi$. LD then recursively updates the continuous time gap with step size $\epsilon>0$ following the equation:
\begin{align} \label{eq:LD-time}
    t^{(n)}  =   t^{(n-1)}  +  \frac{\epsilon}{2}\psi_t^i(t_{i-1} + t^{(n-1)},k^{(n-1)})  + \sqrt{\epsilon}w_n, 
\end{align}
for $n=1,\cdots,N.$ Here, $w_n$ is standard Gaussian noise.
The event mark is updated by sampling $k^{(n)}$ from a categorical distribution defined by $p^i(k\given t_{i-1}+t^{(n-1)})=\frac{\lambda^i(t_{i-1}+t^{(n-1)},k)}{\sum_l \lambda^i(t_{i-1}+t^{(n-1)},l)}$.
After the Langevin sampling, we utilize an additional denoising by Tweedie’s formula \citep{tweedie}, following \cite{smurf2023}:
\begin{align}\label{denoise_step}
    \hat{t} &= t^{(N)} +\sigma_1\psi_t^i(t_{i-1}+t^{(N)}, k^{(N)}).
\end{align}
By sampling $\hat{k}$ from the updated categorical distribution, we obtain the $i$-th event's time and mark as $(\hat{t}, \hat{k})$. 
We then proceed similarly to generate the event's location through $N$-step Langevin Dynamics conditioning on the generated time and mark:
\begin{align} \label{eq:LD-spatial}
    \bm{x}^{(n)} &= \bm{x}^{(n-1)}  +  \frac{\epsilon}{2}\psi_{\bm{x}}^i(\bm{x}^{(n-1)}\given t_{i-1}  +  \hat{t}, \hat{k})  + \sqrt{\epsilon}\bm{z}_n,\\
    \hat{\bm{x}} &= \bm{x}^{(N)} +\bm{\sigma}_2 \psi_{\bm{x}}^i(\bm{x}^{(N)} \given t_{i-1}+\hat{t}, \hat{k}),\label{eq:d-spatial}
\end{align}
where $\bm{x}^{(0)}$ is sampled from an initial uniform distribution and $\bm{z}_n$ is standard multivariate Gaussian noise.
Algorithm \ref{sampling} summarizes the procedure to generate the $i$-th event conditioning on the history $\mathcal{H}_{t_i}$.
\begin{algorithm}
\caption{Sampling $i$-th event given history $\mathcal{H}_{t_i}$}\label{sampling}
\begin{algorithmic}[1]
\STATE \textbf{Target:} generate the $i$-th event $(\hat{t}_i,\hat{\bm{x}}_i, \hat{k}_i)$.
\STATE $(t^{(0)},\bm{x}^{(0)},k^{(0)})\sim \pi$
\FOR{$n=1,2,\cdots,N$}
\STATE $w_n\sim \mathcal{N}(0,1)$
\STATE Update $t^{(n)}$ according to Eq.~\ref{eq:LD-time}
\STATE $k^{(n)}\sim \textnormal{Categorical}(\frac{\lambda^i(t_{i-1}+t^{(n-1)}, k)}{\sum_{l=1}^M \lambda^i(t_{i-1}+t^{(n-1)},l)})$
\ENDFOR\label{euclidendwhile}
\STATE Calculate $\hat{t}$ based on Eq.~\ref{denoise_step}
\STATE $\hat{k} \sim \textnormal{Categorical}(\frac{\lambda^i(t_{i-1}+\hat{t}, k)}{\sum_{l=1}^M \lambda^i(t_{i-1}+\hat{t},l)})$
\FOR{$n=1,2,\cdots,N$}
\STATE $\bm{z}_n\sim \mathcal{N}(\bm{0},\bm{1})$
\STATE Update $\bm{x}^{(n)}$ according to Eq.~\ref{eq:LD-spatial}
\ENDFOR\label{euclidendwhile2}
\STATE Calculate $\hat{\bm{x}}$ based on Eq.~\ref{eq:d-spatial}
\STATE \textbf{return} $\hat{t}_i = \hat{t}+t_{i-1}, \hat{\bm{x}}_i = \hat{\bm{x}},\hat{k}_i=\hat{k}.$
\end{algorithmic}
\end{algorithm}

\textbf{Uncertainty Quantification:} With the generated samples, we can quantify the uncertainty for event predictions. For event time and location, prediction is made by calculating the mean of the samples, while uncertainty is quantified by obtaining the confidence regions from the samples. For event mark, we regard the mode of the samples' mark as the prediction and uncertainty is quantified through the proportion of the predicted mark.

\subsection{Learning Marked Temporal Point Process}
Our proposed method is general and can be easily applied to marked TPP data.
By simply removing the objective for spatial features, we can learn the conditional intensity function $\lambda^i(t,k)$ for marked TPP model through minimizing
\begin{align}\label{eq:smash-tpp}
\widetilde{\mathcal{L}}_{\rm SMASH}^{\rm TPP} (S)= \widetilde{\mathcal{L}}^{\rm D}_{\rm time}(S)+ \alpha  \widetilde{\mathcal{L}}_{\rm mark}(S).
\end{align}
From the learned intensities, event time and mark samples can be jointly generated by the former part of Algorithm~\ref{sampling}.

\citet{smurf2023} focus on marked TPPs rather than STPPs and propose SMURF-THP that applies score matching technique to provide uncertainty quantification. Compared to their work, we jointly model event time and mark by parameterizing an intensity function for each mark, whereas SMURF-THP employs a single intensity function for all event marks, hindering its ability to differentiate event time patterns for different marks. Furthermore, we capture event mark distribution by the joint intensity through a unified model. 
In contrast, SMURF-THP relies on an independent decoder, separated from the intensity function, to predict the mark of the next event, resulting in less accurate modeling of the time-mark dependency.

\section{Experiments}
\label{sec:4}
We present an empirical evaluation of the proposed method by comparing its performance against multiple neural STPP baselines on real-world datasets.  We examine the models' performance on predicting and quantifying uncertainty of the next event given history. As the proposed framework is general and can be directly applied to marked TPPs, we also include a comparison with neural marked TPP models on marked TPP datasets. Further experimental details and dataset statistics can be found in the Appendix.

\textbf{Metrics:} We employ four metrics to evaluate the quality of the generated event samples:

\noindent $\bullet$ \textit{Calibration Score (CS)} measures the uncertainty quantification performance of the samples' time and location. It first computes confidence regions at different confidence levels from the samples, and then calculates the calibration error for each level. The calibration error is determined by the difference between the actual coverage of the region and the desired confidence level. The final metric is defined as the average calibration error across all confidence regions. A lower CS denotes superior performance in quantifying uncertainty. In our experiment, we compute the average error at confidence levels from $0.5$ to $1$ in increments of $0.1$ for STPPs, and from $0.8$ to $1$ in increments of $0.05$ for TPPs. This choice is motivated by that confidence intervals/regions with higher levels are typically more useful.

\noindent $\bullet$ \textit{Mean Absolute Error (MAE)} measures the mean difference between the point prediction and the ground truth of the events' times and locations. We make point predictions by calculating the average time and location of the generated samples.

\noindent $\bullet$ \textit{Mark Prediction Accuracy (Acc)} calculates the accuracy of the event mark predicted by the samples. We designate the mode of the generated samples' mark as the prediction.

\noindent $\bullet$ \textit{Expected Calibration Error (ECE)} evaluates the model's ability to quantify uncertainty for event mark prediction. It measures the discrepancy between the predicted probabilities and the true observed frequencies of outcomes.

\subsection{Uncertainty Quantification for Marked Spatio-Temporal Point Process.}
We first compare SMASH against neural STPP baselines on Spatio-Temporal data, where we evaluate the calibration score and MAE on both event time and location, with accuracy and ECE for event mark prediction.

\textbf{Datasets:} We utilize the following three marked STPP datasets:
(1) \textit{Earthquake} dataset contains the location and time of all earthquakes in Japan from 1990 to 2020 with a magnitude of at least 2.0 from the U.S. Geological Survey \footnote{https://earthquake.usgs.gov/earthquakes/search/}. We partition all earthquakes into three categories: "small", "medium" and "large" based on their magnitude. 
(2) \textit{Crime} dataset comprises reported crime from 2015 to 2020 provided by Atlanta Police Department \footnote{https://www.atlantapd.org}. Events are classified into four types according to the crime type.
(3) \textit{Football} \cite{football} dataset records football event data retrieved from the WyScout Open Access Dataset. Each event signifies an action made by the player, associated with the type of the action.

\textbf{Baselines:} We compare our model against the following five neural STPP methods:
(1) \textit{NJSDE} \citep{SDE_2019} adopts an SDE latent process to model temporal dynamics and utilize mixtures of Gaussian for spatial distribution; 
(2) \textit{NSTPP} \citep{ODESTPP_2021} incorporates an ODE latent process for temporal dynamics and CNFs for spatial distribution.
(3) \textit{NSMPP} \citep{SpecSTPP_2022} utilizes neural networks to construct a non-stationary kernel for modeling the joint conditional intensity.
(4) \textit{DeepSTPP} \citep{DeepSTPP_2022} introduces an RBF kernel-based parametrization for the joint intensity function that supports exact likelihood computation.
(5) \textit{DSTPP} \citep{DSTPP_2023} leverages the diffusion model to capture the complex spatio-temporal dynamics.

\textbf{Overall performance:} Table \ref{tab:stpp} summarizes the results. We can observe that SMASH yields the best performance in terms of CS and MAE for event time, while achieving remarkable improvement in event location modeling. For event mark prediction, SMASH obtains comparable accuracy with DSTPP and the lowest ECE.
The improvements of SMASH can be attributed to two factors: 1) SMASH learns the score function without restricted assumption on the event distribution, whereas NJSDE and DeepSTPP examine higher CS and MAE due to the inaccurate pre-specified parametric form. 2) SMASH optimizes through a normalization-free objective, while NJSDE, NSTPP and NSMPP suffer from inaccurate numerical approximation.

\begin{figure}[htb!]
\centering
\subfigure[Event Time]{\label{all_quantile_crime_t}
\includegraphics[width=0.3\linewidth]{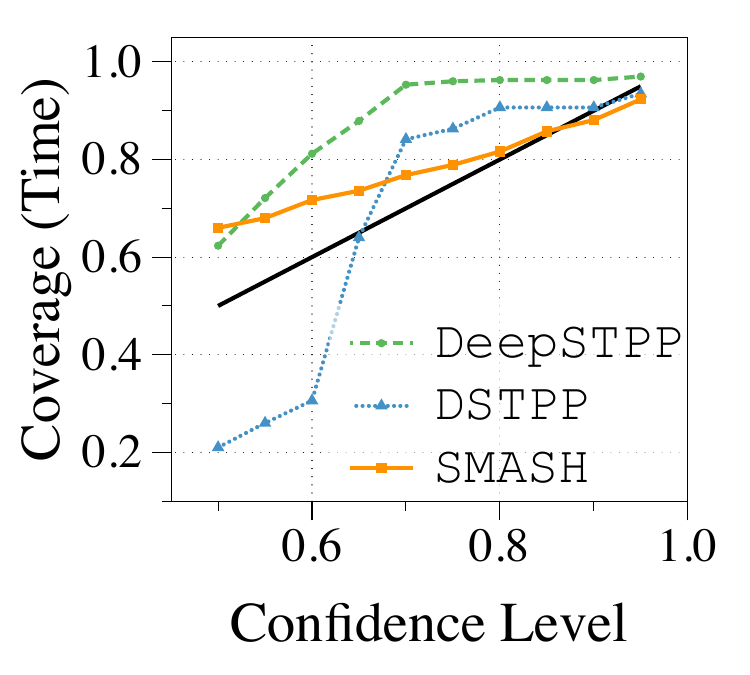}%
}
\subfigure[Event Location]{\label{all_quantile_crime_s}
\includegraphics[width=0.3\linewidth]{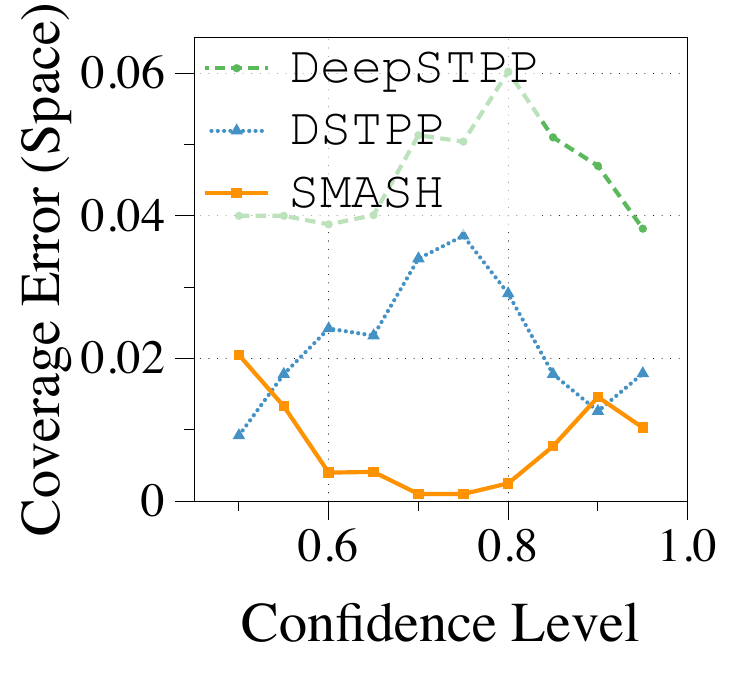}%
}
\vspace{-0.10in}
\caption{Comparison of coverage of different confidence levels on the Crime dataset.}
\label{all_quantile_crime}
\end{figure}

\textbf{Different Confidence Level:} We further compare the coverage of the confidence regions at different levels on the Crime dataset.
Figure~\ref{all_quantile_crime_t} displays the predicted confidence intervals' coverage for event time. The black diagonal represents the ideal coverage. Compared with DeepSTPP and DSTPP, SMASH is much closer to the black line, indicating that the generated confidence regions are more accurate. We can see that DeepSTPP and SMASH experience overconfidence on low confidence levels, while DSTPP tends to be underconfidence. As the level increases, SMASH approaches the correct coverage, whereas DSTPP and DeepSTPP present severe overconfidence.
For event location, we highlight the coverage error of the confidence regions in Figure~\ref{all_quantile_crime_s} to elucidate the distinctions between the models. As shown, SMASH presents the lowest coverage error for most confidence levels, with DeepSTPP being the worst.

\begin{figure}[htb!]
\centering
\subfigure[Event Time]{\label{sample_dis_earthquake_t}
\includegraphics[width=0.3\linewidth]{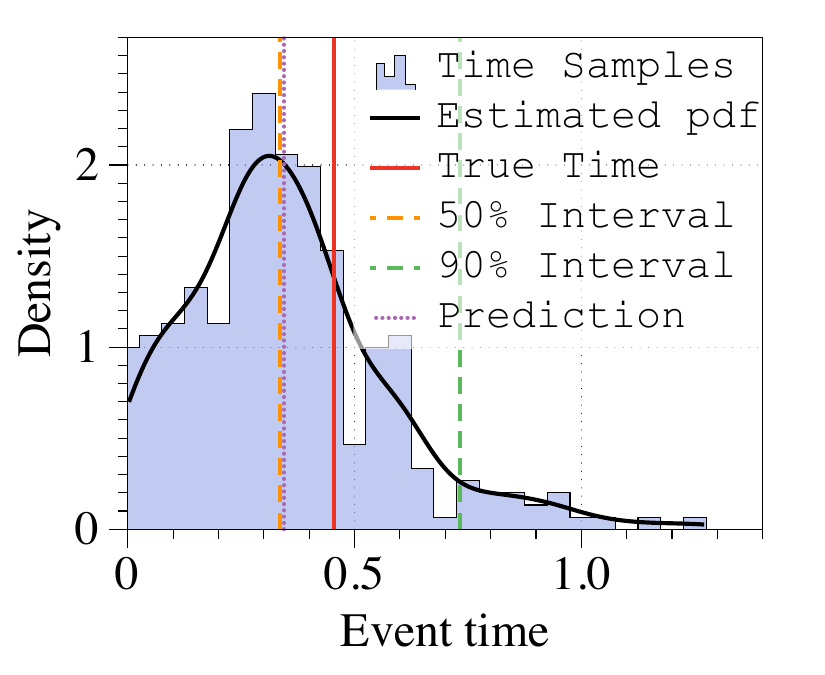}%
}
\subfigure[Event Location]{\label{sample_dis_earthquake_s}
\includegraphics[width=0.3\linewidth]{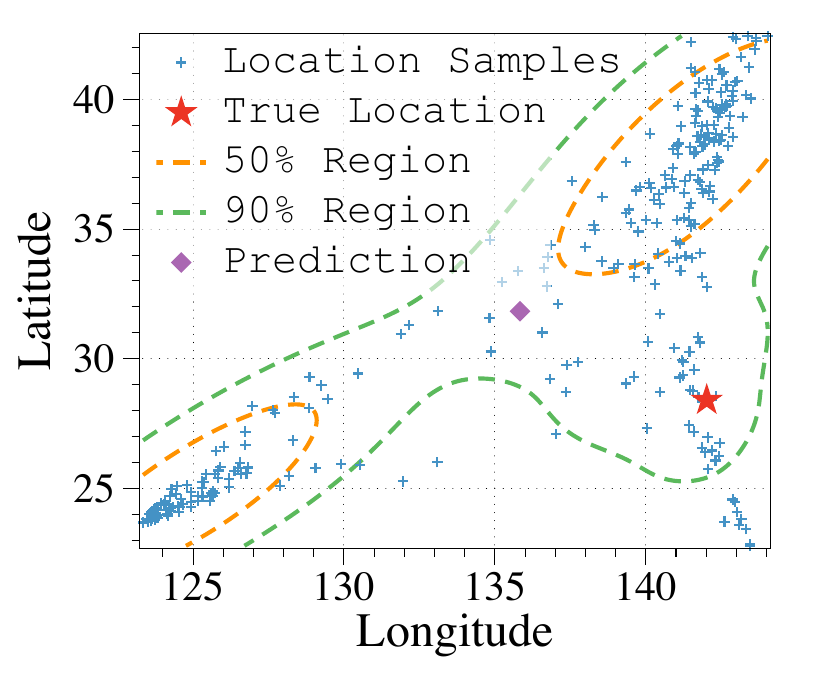}%
}
\vspace{-0.10in}
\caption{Distribution of samples' times and locations generated by SMASH and the ground truth on the Earthquake dataset.}
\label{sample_dis_earthquake}
\end{figure}

\textbf{Sample distribution visualization:} We visualize the distribution of the generated samples for a randomly selected event from the Earthquake dataset. As shown in Figure~\ref{sample_dis_earthquake}, both the samples' times and locations  exhibit a broad spread, which indicates the huge randomness within the event dynamics. Similar distributions is also observed among other baselines. This proves the perspective that single point estimation is not sufficient where uncertainty quantification is needed for the evaluation of the predicted distribution. Furthermore, the samples' locations on the right present a multi-modal distribution, where the point estimation calculated from the average can easily diverge from the actual observation.

\begin{table*}[]
\centering
\caption{Comparison of different methods’ performance on three marked STPP datasets in terms of Calibration Score (CS) and Mean Absolute Error (MAE) for event time and location, Accuracy (Acc) and Expected Calibration Error (ECE) for event mark.}
\label{tab:stpp}
\vspace{0.05in}
\resizebox{\textwidth}{!}{%
\begin{tabular}{clcccccc}
\toprule
Datasets                                         & Methods  & $\text{CS}_{\rm time}(\downarrow)(\%)$ & $\text{MAE}_{\rm time}(\downarrow)(\%)$ & $\text{CS}_{\rm space}(\downarrow)(\%)$ & $\text{MAE}_{\rm space}(\downarrow)(\%)$ & $\text{Acc}(\uparrow)(\%)$    & $\text{ECE}(\downarrow)(\%)$      \\ \hline
\multicolumn{1}{c|}{\multirow{6}{*}{Earthquake}} & NJSDE    & $16.23\scriptstyle{\pm 0.55}$          & $0.41\scriptstyle{\pm 0.08}$            & $8.35\scriptstyle{\pm 0.35}$            & $8.05\scriptstyle{\pm 0.04}$             & $90.03\scriptstyle{\pm 0.05}$ & $11.23\scriptstyle{\pm 0.22}$     \\
\multicolumn{1}{c|}{}                            & NSTPP    & $14.66\scriptstyle{\pm 0.89}$          & $0.45\scriptstyle{\pm 0.12}$            & $7.33\scriptstyle{\pm 0.43}$            & $7.45\scriptstyle{\pm 0.05}$             & $90.13\scriptstyle{\pm 0.03}$ & $12.21\scriptstyle{\pm 0.12}$     \\
\multicolumn{1}{c|}{}                            & NSMPP    & $25.30\scriptstyle{\pm 0.25}$          & $0.69\scriptstyle{\pm 0.03}$            & $8.66\scriptstyle{\pm 0.14}$            & $7.99\scriptstyle{\pm 0.03}$             & $88.20\scriptstyle{\pm 0.03}$ & $10.47\scriptstyle{\pm 0.03}$     \\
\multicolumn{1}{c|}{}                            & DeepSTPP & $22.77\scriptstyle{\pm 1.32}$          & $0.38\scriptstyle{\pm 0.01}$            & $7.31\scriptstyle{\pm 0.25}$            & $7.55\scriptstyle{\pm 0.10}$             & $90.11\scriptstyle{\pm 0.09}$ & $21.20\scriptstyle{\pm 1.50}$     \\
\multicolumn{1}{c|}{}                            & DSTPP    & $14.80\scriptstyle{\pm 0.67}$          & $0.35\scriptstyle{\pm 0.01}$            & $\bm{5.81}\scriptstyle{\pm 0.56}$       & $7.42\scriptstyle{\pm 0.08}$             & $90.30\scriptstyle{\pm 0.01}$ & $7.30\scriptstyle{\pm 1.45}$      \\ \cmidrule{2-8} 
\multicolumn{1}{c|}{}                            & SMASH    & $\bm{3.53}\scriptstyle{\pm 0.55}$      & $\bm{0.29}\scriptstyle{\pm 0.01}$       & $6.95\scriptstyle{\pm 0.21}$            & $\bm{7.37}\scriptstyle{\pm 0.06}$        & $90.30\scriptstyle{\pm 0.01}$ & $\bm{4.85}\scriptstyle{\pm 0.09}$ \\ \midrule
\multicolumn{1}{c|}{\multirow{6}{*}{Crime}}      & NJSDE    & $17.33\scriptstyle{\pm 0.25}$          & $1.72\scriptstyle{\pm 0.21}$            & $7.89\scriptstyle{\pm 0.70}$              & $0.083\scriptstyle{\pm 0.20}$            & $36.26\scriptstyle{\pm 0.04}$ & $13.42\scriptstyle{\pm 0.65}$     \\
\multicolumn{1}{c|}{}                            & NSTPP    & $17.26\scriptstyle{\pm 0.69}$          & $1.41\scriptstyle{\pm 0.14}$            & $7.03\scriptstyle{\pm 0.98}$            & $0.065\scriptstyle{\pm 0.35}$            & $37.03\scriptstyle{\pm 0.04}$ & $14.20\scriptstyle{\pm 0.28}$     \\
\multicolumn{1}{c|}{}                            & NSMPP    & $16.09\scriptstyle{\pm 0.25}$          & $1.63\scriptstyle{\pm 0.12}$            & $9.15\scriptstyle{\pm 0.08}$            & $0.058\scriptstyle{\pm 0.017}$           & $36.30\scriptstyle{\pm 0.03}$ & $15.47\scriptstyle{\pm 0.23}$     \\
\multicolumn{1}{c|}{}                            & DeepSTPP & $16.64\scriptstyle{\pm 0.31}$          & $1.16\scriptstyle{\pm 0.02}$            & $5.04\scriptstyle{\pm 0.79}$            & $0.058\scriptstyle{\pm 0.010}$           & $36.25\scriptstyle{\pm 0.03}$ & $15.14\scriptstyle{\pm 0.62}$     \\
\multicolumn{1}{c|}{}                            & DSTPP    & $14.50\scriptstyle{\pm 1.81}$          & $1.43\scriptstyle{\pm 0.13}$            & $1.92\scriptstyle{\pm 1.48}$            & $0.057\scriptstyle{\pm 0.012}$           & $39.33\scriptstyle{\pm 0.93}$ & $14.65\scriptstyle{\pm 1.53}$     \\ \cmidrule{2-8}  
\multicolumn{1}{c|}{}                            & SMASH    & $\bm{6.93}\scriptstyle{\pm 0.36}$      & $1.16\scriptstyle{\pm 0.06}$            & $\bm{1.08}\scriptstyle{\pm 0.12}$       & $0.057\scriptstyle{\pm 0.008}$           & $39.40\scriptstyle{\pm 1.56}$ & $\bm{11.36}\scriptstyle{\pm 0.78}$     \\ \midrule
\multicolumn{1}{c|}{\multirow{6}{*}{Football}}   & NJSDE    & $19.21\scriptstyle{\pm 0.55}$          & $4.85\scriptstyle{\pm 0.25}$            & $11.32\scriptstyle{\pm 0.75}$           & $36.66\scriptstyle{\pm 0.20}$            & $65.58\scriptstyle{\pm 0.63}$ & $9.53\scriptstyle{\pm 0.21}$      \\
\multicolumn{1}{c|}{}                            & NSTPP    & $18.36\scriptstyle{\pm 0.73}$          & $4.66\scriptstyle{\pm 0.30}$            & $7.23\scriptstyle{\pm 0.98}$            & $27.57\scriptstyle{\pm 0.35}$            & $66.33\scriptstyle{\pm 0.84}$ & $10.68\scriptstyle{\pm 0.43}$     \\
\multicolumn{1}{c|}{}                            & NSMPP    & $19.02\scriptstyle{\pm 1.19}$          & $4.85\scriptstyle{\pm 0.41}$            & $8.62\scriptstyle{\pm 0.57}$            & $30.65\scriptstyle{\pm 0.65}$            & $65.80\scriptstyle{\pm 0.21}$ & $9.37\scriptstyle{\pm 0.22}$      \\
\multicolumn{1}{c|}{}                            & DeepSTPP & $19.54\scriptstyle{\pm 1.25}$          & $4.24\scriptstyle{\pm 0.36}$            & $9.24\scriptstyle{\pm 0.18}$            & $32.54\scriptstyle{\pm 0.79}$            & $65.92\scriptstyle{\pm 0.05}$ & $12.10\scriptstyle{\pm 0.08}$     \\
\multicolumn{1}{c|}{}                            & DSTPP    & $14.53\scriptstyle{\pm 4.22}$          & $3.76\scriptstyle{\pm 0.24}$            & $2.79\scriptstyle{\pm 0.81}$            & $\bm{19.41}\scriptstyle{\pm 0.07}$       & $68.62\scriptstyle{\pm 0.86}$ & $4.64\scriptstyle{\pm 0.83}$      \\ \cmidrule{2-8} 
\multicolumn{1}{c|}{}                            & SMASH    & $\bm{6.38}\scriptstyle{\pm 0.54}$      & $\bm{3.40}\scriptstyle{\pm 0.08}$       & $\bm{2.52}\scriptstyle{\pm 0.34}$       & $20.35\scriptstyle{\pm 0.13}$            & $68.38\scriptstyle{\pm 1.03}$ & $\bm{3.23}\scriptstyle{\pm 1.68}$ \\ \bottomrule
\end{tabular}%
}
\end{table*}

\subsection{Uncertainty Quantification for Marked Temporal Point Process.}
As SMASH provides a general framework, we can also model marked TPP data using score-based objectives by simply removing the loss term of the spatial location. In this subsection, we compare SMASH with neural TPP models on four real-world marked TPP data.

\textbf{Datasets:} We utilize the following four real-world datasets:
(1) \textit{StackOverflow} \cite{so_data}. This dataset contains sequences of $6,633$ users' receipt of awards from a question answering website over a two-year period. Events are marked according to the type of the award, with a total of 21 distinct types.
(2) \textit{Retweet} \citep{retweet_data} dataset comprises $24,000$ sequences of tweets, where each sequence begins with the original tweet at time 0. Subsequent events represent retweets by other users, which are classified into three categories based on their follower counts.
(3) \textit{MIMIC-II} \citep{rmtpp2016} dataset is a comprehensive collection of clinical data from patients admitted to intensive care units (ICUs) over a seven-year period. We select a subset of $650$ patients and construct sequences from their visit time and diagnosis codes.
(4) \textit{Financial Transactions} \citep{rmtpp2016} dataset records a total of $0.7$ million transaction actions for a stock from the New York Stock Exchange. We partition the long single sequence of transactions into $2,000$ subsequences for evaluation. Each event is labeled with the transaction time and the action that was taken: buy or sell.

\textbf{Baselines:} We compare our model against the following five existing methods:
(1) \textit{NHP} \citep{nhp2017} designs a continuous-time LSTM to model the evolution of the intensity function by updating the latent state; 
(2) \textit{NCE-TPP} \citep{ncetpp2020} utilizes noise-contrastive estimation on MTPPs to bypass likelihood computation.
(3) \textit{SAHP} \citep{sahp2020} introduces a time-shifted positional encoding and employs self-attention to model the intensity function.
(4) \textit{THP} \citep{thp2020} leverages the Transformer architecture to capture long-term dependencies in history and parameterizes the intensity function through a tailored continuous formulation. 
(5) \textit{SMURF-THP} \citep{smurf2023} develops a score matching-based objective to train the THP model and provides uncertainty quantification for predicted arrival time.

\textbf{Overall performance:} Table~\ref{tab:main} summarizes the results.
We can see that SMASH outperforms other baselines by noticeable margins in terms of CS, and it achieves the lowest MAE on three datasets.
Although SMURF-THP also leverages score matching for modeling marked TPPs, SMASH better captures the time-mark dependencies by modeling the joint intensity function and constructing a unified model.
Additionally, SMASH also improves the ECE while exhibiting comparable accuracy with other baselines. We attribute it to the fact that we perturb event time with noise to use denoising score matching, which implicitly induces regularization for event mark prediction.

\begin{table*}[htb!]
\centering
\caption{Comparison of different methods' performance on four marked TPP datasets in terms of Calibration Score (CS), Mean Absolute Error (MAE) for event time, Accuracy (Acc) and Expected Calibration Error (ECE) for event mark.}
\vspace{0.05in}
\label{tab:main}
\resizebox{\textwidth}{!}{%
\begin{tabular}{lcccccccc}
\toprule
\multicolumn{1}{c}{}            & \multicolumn{4}{c}{StackOverflow}                                                                                           & \multicolumn{4}{c}{Retweet}                                                                                                   \\ \cmidrule(r){2-5} \cmidrule(r){6-9} 
\multicolumn{1}{c}{Methods}     & $\mathrm{CS}(\%)(\downarrow)$ & $\mathrm{MAE}(\downarrow)$  & $\mathrm{Acc}(\%)(\uparrow)$  & $\mathrm{ECE}(\%)(\downarrow)$   & $\mathrm{CS}(\%)(\downarrow)$ & $\mathrm{MAE}(\downarrow)$  & $\mathrm{Acc}(\%)(\uparrow)$  & $\mathrm{ECE}(\%)(\downarrow)$ \\ \hline
$\mathrm{NHP}$                  & $1.18\scriptstyle{\pm 0.21}$  & $0.72\scriptstyle{\pm 0.01}$ & $46.26\scriptstyle{\pm 0.02}$ & $5.22\scriptstyle{\pm 0.06}$ & $3.78\scriptstyle{\pm 0.28}$  & $1.63\scriptstyle{\pm 0.02}$ & $60.69\scriptstyle{\pm 0.11}$ & $2.63\scriptstyle{\pm 0.22}$   \\
$\mathrm{NCE}$-$\mathrm{TPP}$   & $1.35\scriptstyle{\pm 0.28}$  & $0.67\scriptstyle{\pm 0.01}$ & $46.02\scriptstyle{\pm 0.04}$ & $5.56\scriptstyle{\pm 0.13}$ & $2.73\scriptstyle{\pm 0.35}$  & $1.46\scriptstyle{\pm 0.04}$ & $60.01\scriptstyle{\pm 0.23}$ & $2.42\scriptstyle{\pm 0.24}$   \\
$\mathrm{SAHP}$                 & $0.85\scriptstyle{\pm 0.21}$  & $0.67\scriptstyle{\pm 0.01}$ & $46.15\scriptstyle{\pm 0.03}$ & $5.13\scriptstyle{\pm 0.08}$ & $1.71\scriptstyle{\pm 0.15}$  & $1.10\scriptstyle{\pm 0.02}$ & $60.65\scriptstyle{\pm 0.13}$ & $2.57\scriptstyle{\pm 0.18}$   \\
$\mathrm{THP}$                  & $1.36\scriptstyle{\pm 0.40}$  & $0.63\scriptstyle{\pm 0.01}$ & $46.50\scriptstyle{\pm 0.02}$ & $5.42\scriptstyle{\pm 0.10}$ & $3.43\scriptstyle{\pm 0.51}$  & $1.59\scriptstyle{\pm 0.04}$ & $60.82\scriptstyle{\pm 0.06}$ & $2.96\scriptstyle{\pm 0.33}$   \\
$\mathrm{SMURF}$-$\mathrm{THP}$ & $0.34\scriptstyle{\pm 0.04}$  & $0.64\scriptstyle{\pm 0.01}$ & $46.26\scriptstyle{\pm 0.05}$ & $5.41\scriptstyle{\pm 0.04}$ & $0.35\scriptstyle{\pm 0.04}$  & $0.99\scriptstyle{\pm 0.01}$ & $60.80\scriptstyle{\pm 0.08}$ & $2.45\scriptstyle{\pm 0.11}$   \\
\midrule
$\mathrm{SMASH}$ & $\mathbf{0.29}\scriptstyle{\pm 0.12}$  & $0.63\scriptstyle{\pm 0.01}$ & $46.26\scriptstyle{\pm 0.14}$ & $\mathbf{3.89}\scriptstyle{\pm 0.03}$ & $\mathbf{0.27}\scriptstyle{\pm 0.09}$ & $0.96\scriptstyle{\pm 0.01}$ & $60.80\scriptstyle{\pm 0.06}$ & $\mathbf{2.23}\scriptstyle{\pm 0.05}$   \\ \midrule
\multicolumn{1}{c}{}            & \multicolumn{4}{c}{Financial}                                                                                               & \multicolumn{4}{c}{MIMIC}                                                                                                     \\ \cmidrule(r){2-5} \cmidrule(r){6-9}  
\multicolumn{1}{c}{Methods}     & $\mathrm{CS}(\%)(\downarrow)$ & $\mathrm{MAE}(\downarrow)$  & $\mathrm{Acc}(\%)(\uparrow)$  & $\mathrm{ECE}(\%)(\downarrow)$   & $\mathrm{CS}(\%)(\downarrow)$ & $\mathrm{MAE}(\downarrow)$  & $\mathrm{Acc}(\%)(\uparrow)$  & $\mathrm{ECE}(\%)(\downarrow)$ \\ \hline
$\mathrm{NHP}$                  & $1.66\scriptstyle{\pm 0.21}$  & $1.96\scriptstyle{\pm 0.05}$ & $60.39\scriptstyle{\pm 0.25}$ & $4.15\scriptstyle{\pm 0.13}$ & $1.43\scriptstyle{\pm 0.10}$  & $0.99\scriptstyle{\pm 0.01}$ & $83.10\scriptstyle{\pm 0.91}$ & $15.80\scriptstyle{\pm 0.63}$  \\
$\mathrm{NCE}$-$\mathrm{TPP}$   & $1.64\scriptstyle{\pm 0.27}$  & $2.30\scriptstyle{\pm 0.16}$ & $60.12\scriptstyle{\pm 0.08}$ & $4.38\scriptstyle{\pm 0.35}$ & $1.36\scriptstyle{\pm 0.68}$  & $1.13\scriptstyle{\pm 0.01}$ & $83.17\scriptstyle{\pm 0.67}$ & $14.62\scriptstyle{\pm 1.09}$  \\
$\mathrm{SAHP}$                 & $1.38\scriptstyle{\pm 0.30}$  & $1.61\scriptstyle{\pm 0.05}$ & $60.83\scriptstyle{\pm 0.12}$ & $3.85\scriptstyle{\pm 0.86}$ & $1.36\scriptstyle{\pm 0.46}$  & $0.87\scriptstyle{\pm 0.01}$ & $82.10\scriptstyle{\pm 0.87}$ & $21.56\scriptstyle{\pm 0.99}$  \\
$\mathrm{THP}$                  & $1.54\scriptstyle{\pm 0.01}$  & $1.89\scriptstyle{\pm 0.01}$ & $60.84\scriptstyle{\pm 0.30}$ & $3.48\scriptstyle{\pm 0.22}$ & $1.20\scriptstyle{\pm 0.37}$  & $1.09\scriptstyle{\pm 0.01}$ & $83.73\scriptstyle{\pm 0.05}$ & $13.35\scriptstyle{\pm 0.81}$  \\
$\mathrm{SMURF}$-$\mathrm{THP}$ & $1.28\scriptstyle{\pm 0.06}$  & $1.40\scriptstyle{\pm 0.01}$ & $60.85\scriptstyle{\pm 0.38}$      & $3.71\scriptstyle{\pm 0.15}$ & $1.14\scriptstyle{\pm 0.23}$  & $0.87\scriptstyle{\pm 0.01}$ & $83.72\scriptstyle{\pm 0.48}$ & $15.65\scriptstyle{\pm 0.85}$                              \\
\midrule
$\mathrm{SMASH}$ & $\mathbf{0.81}\scriptstyle{\pm 0.21}$  & $1.42\scriptstyle{\pm 0.01}$ & $60.95\scriptstyle{\pm 0.37}$ & $\mathbf{2.30}\scriptstyle{\pm 0.71}$ & $\mathbf{0.85}\scriptstyle{\pm 0.38}$  & $0.87\scriptstyle{\pm 0.02}$ & $83.72\scriptstyle{\pm 0.13}$ & $\mathbf{12.23}\scriptstyle{\pm 0.70}$  \\ \bottomrule

\end{tabular}%
}
\end{table*}

\subsection{Ablation Study}
\begin{figure}[htb!]
\centering
\subfigure[Event Time]{
\includegraphics[width=0.3\linewidth]{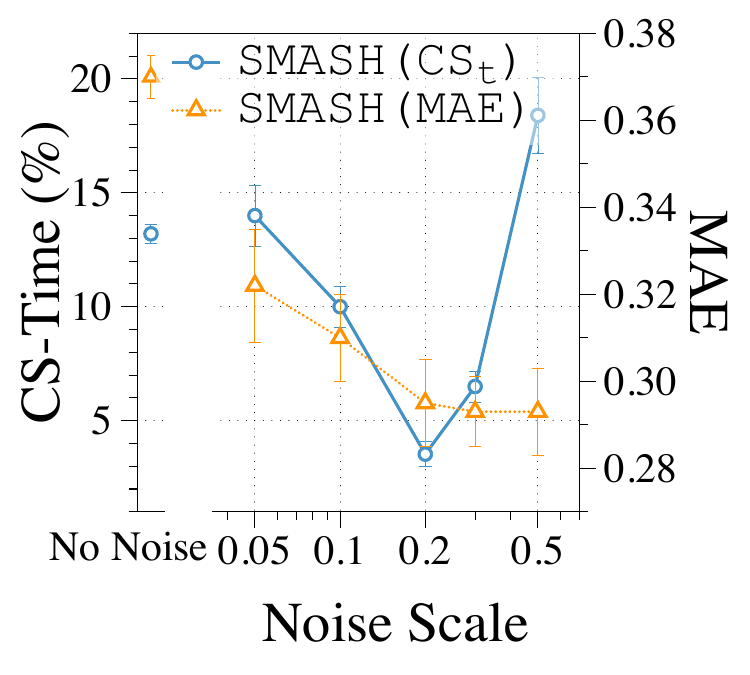}%
}
\subfigure[Event Location]{
\includegraphics[width=0.3\linewidth]{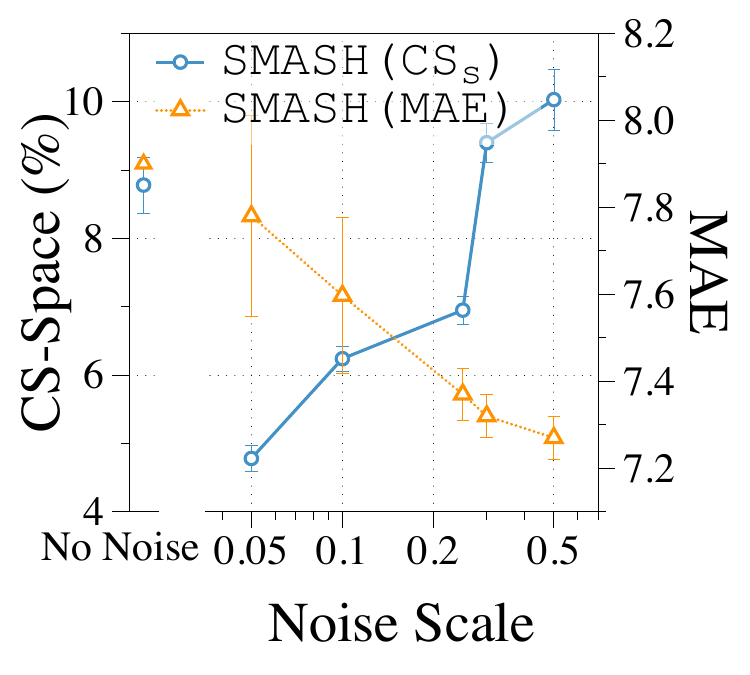}%
}
\vspace{-0.10in}
\caption{Performance of SMASH with different noise scales on the Earthquake dataset.}
\label{fig:earth_noise}
\end{figure}

\textbf{Denoising:} SMASH perturbs data points with Gaussian noise and employs denoising score matching to achieve better stability and computational efficiency.
We investigate the effects of noise scale $\sigma$ by adding different amounts of noise to the data, including training without noise added.
Results tested on the Earthquake dataset are presented in Figure \ref{fig:earth_noise}.
The figures suggest that adding perturbations effectively improves performance on both event time and location when a suitable noise scale is chosen.
The calibration score of event time first decreases and then increases as the noise grows, while the calibration score of event location manifests a continuous increase.
This suggests that a small noise is sufficient to cover the low-density regions in spatial distribution, while the event time requires a larger noise magnitude.
Notably, the MAE of both event time and location decrease and tend to converge as the noise increase, implying that larger noise can bring smaller mean bias.

\begin{figure}[htb!]
\centering
\subfigure[Calibration Score]{
    \includegraphics[width=0.33\linewidth]{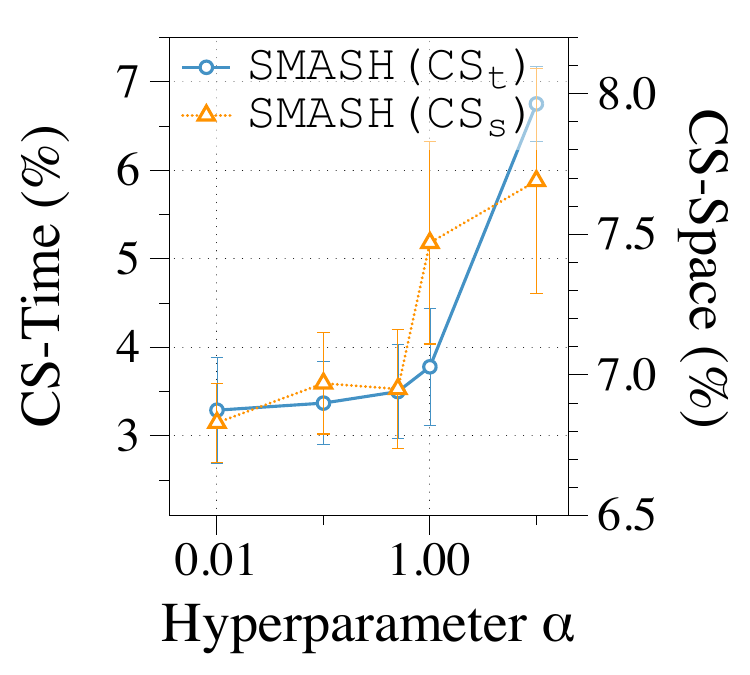}%
}
\subfigure[Event Mark]{  
    \includegraphics[width=0.33\linewidth]{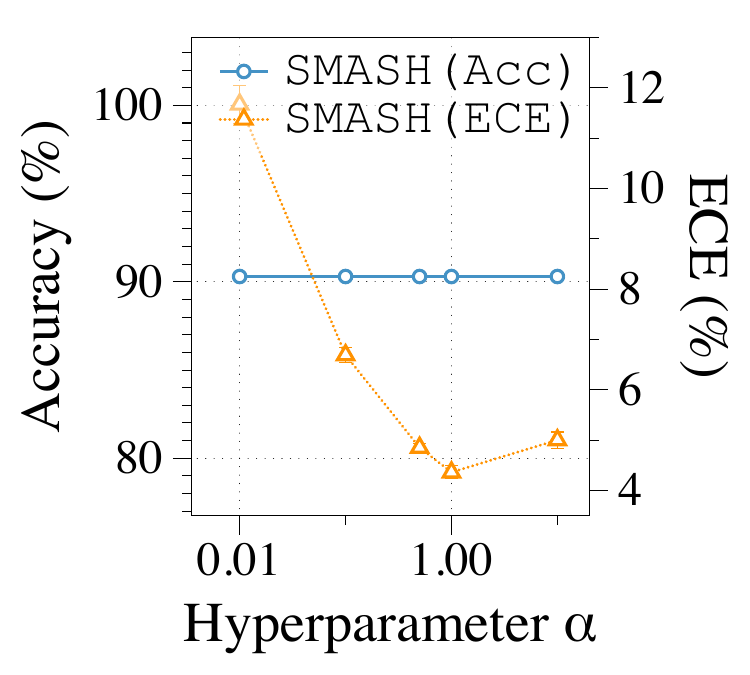}%
}
\vspace{-0.10in}
\caption{Sensitivity of the hyperparameter $\alpha$ in the training objective to model's performance on the Earthquake dataset.}
\label{fig:earth_loss}
\end{figure}

\textbf{Hyperparameter $\alpha$.} We investigate our model's sensitivity to the hyperparameter $\alpha$ in the training objectives on the Earthquake dataset.
As depicted in Figure \ref{fig:earth_loss}, the calibration scores of both event time and location follow an increasing trend as $\alpha$ increases, while the ECE for event mark shows improvement.
This observation aligns with our understanding of $\alpha$ as a scaler of the cross-entropy objective for event mark modeling, and increasing it places more emphasis on fitting event mark distribution.
The slight increase in ECE at $\alpha$ being 10 may hint at overfitting within the mark distribution.

\section{Discussions}
\label{sec:5}
We remark that our work bears similarity to that of \citet{DSTPP_2023}, which applies a denoising diffusion probabilistic model to STPPs.
Their approach introduces multiple scales of noise to perturb the event distribution and learns all perturbed distributions throughout the process. On the contrary, we only add the noise once and learn the corresponding perturbed distribution, which largely simplifies the framework.
Notably, our experimental results indicate that a complex diffusion process is not necessary for STPP modeling, which may be attributable to the lower data dimension compared to the original high-dimensional image use case.

In parallel, several studies on temporal point process (TPP) explore non-likelihood-based objectives to circumvent the computationally challenging integral calculation. For instance, \citet{DIS2017} apply a discriminative learning method to estimate the model's parameters. However, it does not support flexible sampling and uncertainty quantification due to the absence of an explicit intensity function.
TPPRL \citep{TPPRL2018} employ reinforcement learning (RL) for learning a policy to generate events in the setting of temporal point process (TPP), which can not be trivially extended to marked TPP/STPP. 
RLPP \citep{RLPP2018} apply RL to marked TPP under an overly restrictive assumption of exponential intensity functions, which limits the ability to capture complex point processes in real-world applications.
INITIATOR \citep{initiator2018} and NCE-TPP \citep{ncetpp2020} adopt noise-contrastive estimations to model marked TPP,  while these two methods still need to deal with the intractable integral due to the likelihood-based approach for the training of noise generation network. As NCE-TPP has been shown to outperform INITIATOR by the authors, we include NCE-TPP in our set of baseline comparisons.

\vspace{0.10in}

\section{Conclusion}

In this work, we present SMASH, a Score-MAtching based pSeudolikeliHood estimator for learning marked STPPs. We begin by decomposing the intensity function to separate the spatial features, thereby circumventing the intractable spatial integral that arises when applying the score-matching technique. Then we derive the score-matching objectives for the conditional likelihood of event time and location, and integrate the conditional likelihood of event mark as part of the objective. 
Uncertainty quantification for event time, location and mark is obtained through flexible sampling using Langevin dynamics and the learned score function. We conduct experiments on various real-world datasets to illustrate that SMASH achieves superior performance under both marked STPPs and TPPs settings.

\bibliographystyle{apalike}
\bibliography{main_arxiv} 

\newpage
\onecolumn
\appendix
\input{appendix}

\end{document}

%% file: appendix.tex
\title{Supplementary Material For Score Matching-based Pseudolikelihood Estimation of
Neural Marked Spatio-Temporal Point Process}

\section{Model Architecture}

We utilize the self-attention mechanism to encode both marked STPP and TPP data, as validated by previous research \citep{thp2020, sahp2020, DeepSTPP_2022, DSTPP_2023}. This mechanism captures long-term dependencies by assigning an attention weight between any two events, with higher weights signifying stronger dependencies between those events. We adopt the approach of \cite{DSTPP_2023} and utilize the Transformer Hawkes Process model \citep{thp2020} to parameterize our intensity functions. Here we detail the model for marked STPP data.

Initially, each event in the sequence is encoded by summing the temporal encoding $\mathbf{C}_t$, spatial encoding $\mathbf{C}_{\bm{x}}$ and event mark embedding $\mathbf{C}_k$ \citep{DSTPP_2023, thp2020}, where $\mathbf{C}_t, \mathbf{C}_{\bm{x}}, \mathbf{C}_k \in \mathbb{R}^{L\times d}$ with $d$ being the dimension of embedding. Through this process, the sequence $S$ is encoded as $\mathbf{C}=\mathbf{C}_t+\mathbf{C}_{\bm{x}}+\mathbf{C}_k$. We then pass $\mathbf{C}$ along with each embedding $\mathbf{C}_t, \mathbf{C}_{\bm{x}}, \mathbf{C}_k$ through the self-attention module. Taking $\mathbf{C}$ as the example, the attention o./utput $\mathbf{A}$ is computed as:
\begin{gather*}
\mathbf{A}={\mathrm{Softmax}}\biggl(\frac{\mathbf{Q}^{\top}\mathbf{K}}{\sqrt{d_k}}\biggr)\mathbf{V}, \\
\mathbf{Q}=\mathbf{CW}^{Q},\quad \mathbf{K}=\mathbf{CW}^{K}, \quad \mathbf{V}=\mathbf{CW}^{V}.
\end{gather*}

The matrices $\mathbf{W}^{Q},\mathbf{W}^{K} \in \mathbb{R}^{d\times d_k}$ and $\mathbf{W}^{V} \in \mathbb{R}^{d\times d_v}$ serve as weights for linear transformations that transform $\mathbf{C}$ into query, key, and value, respectively. To enhance model capacity, \cite{attention} suggest using multi-head self-attention. This involves inducing multiple sets of weights $\{\mathbf{W}_h^{Q},\mathbf{W}_h^{K},\mathbf{W}_h^{V}\}_{h=1}^H$ and computing different attention outputs $\{A_h\}_{h=1}^H$. The final attention output for the event sequence is obtained by concatenating $\{A_h\}_{h=1}^H$ and aggregating them with $\mathbf{W}^{O}\in\mathbb{R}^{(d_v*H)\times d}$:
\begin{equation*}
\mathbf{A}=\mathrm{Concat}(\mathbf{A}_1,...,\mathbf{A}_H)\mathbf{W}^{O}.
\end{equation*}
The attention output $\mathbf{A}$ is then processed through a position-wise feed-forward neural network (FFN) to obtain the hidden representations:
\begin{gather*}
\mathbf{H} = \mathrm{ReLU}(\mathbf{AW}_1^\mathrm{FFN}+\mathbf{b_1})\mathbf{W}_2^\mathrm{FFN} +\mathbf{b_2},\\
\mathbf{h}(i) = \mathbf{H}(i,:).
\end{gather*}
In this context, $\mathbf{h}(i)$ encodes the $i$-th event and all past events up to time $t_i$. We incorporate future masks during the computation of attention to prevent learning from future events. We stack multiple self-attention modules and FFNs to construct a larger model that can capture high-level dependencies.

Following the above computation, we obtain $\mathbf{h}(i), \mathbf{h}_t(i), \mathbf{h}_{\bm{x}}(i), \mathbf{h}_k(i)$ as the encoding of different aspects of the event history.
We then parametrize the intensity functions $\lambda^i(t,k)$ given history $\mathcal{H}_{t_i}$ using multiple layers of network, where each layer follows:
\begin{align*}
\mathbf{h}^i_{tk} = \sigma (\mathbf{W}_t t + \mathbf{b}_t + \mathbf{W}_h \mathbf{h}(i) +\mathbf{b}_h + \mathbf{W}_{tk} (\mathbf{h}_t(i) + \mathbf{h}_k(i)) +\mathbf{b}_{tk}).
\end{align*}
Here, $\mathbf{W}_t\in \mathbb{R}^{d\times 1}, \mathbf{W}_h, \mathbf{W}_{tk}\in \mathbb{R}^{d\times d},  \mathbf{b}_t, \mathbf{b}_h, \mathbf{b}_{tk} \in \mathbb{R}^{d}$ are trainable parameters. $\sigma$ denotes the RELU activation function.
We stack three such layers and pass the output to another FFN, with the softplus activation in the last layer:
\begin{align*}
\lambda^i(t,k) = \mathrm{Softplus} (\mathrm{FFN}(\mathbf{h}^i_{tk})).
\end{align*}
For score function of event location, we employ the Co-attention Denoising Network (CDN) proposed by \citet{DSTPP_2023}:
\begin{align*}
\psi^i(\bm{x}\given t,k) = \mathrm{CDN} (\mathbf{h}(i), \mathbf{h}_t(i), \mathbf{h}_{\bm{x}}(i), \mathbf{h}_k(i)).
\end{align*}



\section{Experiment Detail}
\label{app:exp}

\subsection{Training Detail}
\textbf{Dataset Preprocessing: } (1) Earthquake: We adopt the same preprocessing procedure as in \citet{ODESTPP_2021}, with the exception that we exclude earthquakes with magnitudes below 2.0. (2) Crime: We select crime events from 2015 to 2020 at Atlanta with the following crime types: "Burglary", "Agg Assault", "Robbery" and "Homicide". The occurrence time of the crime serves as the event time, with longitude and latitude representing spatial features. (3) Football: We follow the same preprocessing as \citet{football}. (4) Four TPP datasets: We adopt the same data preprocessing as those used by \cite{rmtpp2016} and \cite{nhp2017}. For additional details and downloadable links, please refer to the aforementioned papers. Table~\ref{dataset} summarizes the statistics of the seven datasets used in the experiments. We randomly split all datasets to train/test/valid by the proportion of 0.8/0.1/0.1.  As the datasets' scales differ, we employ normalization and log-normalization techniques. Specifically, we log-normalize event time by $\frac{\log(t)-\mathrm{Mean}(\log(t))}{\mathrm{Var}(\log(t))}$ and apply standard normalization for event location. We rescale back the generated samples before evaluation.

\textbf{Hyperparameters:} In STPP experiments, we employ the same backbone architecture as DSTPP to ensure a fair comparison and maintain default hyperparameters for other baselines. In TPP experiments, we employ the same backbone architecture as SMURF-THP. We note that while the original NCE-TPP uses LSTM as the backbone, we implement it using the same Transformer backbone as ours. A detailed hyperparameter breakdown for the adopted backbone architecture can be found in Table~\ref{hyper:backbone}. During training, we use the Adam optimizer and train all models for 150 epochs on an NVIDIA Tesla V100 GPU.
Hyperparameters specific to our method were fine-tuned via grid search and are detailed in Table~\ref{hyper:method}. The number of perturbations and samples are fixed at 300 for STPPs and 100 for TPPs; increasing these may improve the models' performance. The two numbers of noise scale on marked STPP datasets signify the noise scale on event time and location.

\begin{table}[h]
\centering
\caption{Summary of backbone architecture hyperparameters.}
\label{hyper:backbone}
\begin{tabular}{ccccccccc}
\toprule
Dataset       & \#head & \#layer & $d_\mathrm{model}$ & $d_k=d_v$ & $d_{\mathrm{hidden}}$ & dropout & batch & learning rate  \\ \midrule
Earthquake & 4     & 4      & 16          & 16        & 64          & 0.1     & 32     & 1e-3   \\
Crime      & 4     & 4      & 64          & 16        & 256          & 0.1     & 32    & 1e-3   \\
Football     & 4     & 4      & 32         & 16        & 128         & 0.1     & 4     & 1e-3   \\ 
StackOverflow & 4     & 4      & 64          & 16        & 256          & 0.1     & 4     & 1e-3   \\
Retweet       & 3     & 3      & 64          & 16        & 256          & 0.1     & 16    & 5e-3   \\
Financial     & 6     & 6      & 128         & 64        & 2048         & 0.1     & 1     & 1e-4   \\ 
MIMIC-II      & 3     & 3      & 64          & 16        & 256          & 0.1     & 1     & 1e-4   \\
\bottomrule
\end{tabular}%
\vspace{-2ex}
\end{table}

\begin{table}[h]
\centering
\caption{Summary of hyperparameters for the method.}
\label{hyper:method}
\begin{tabular}{ccccc}
\toprule
Dataset       & Loss weight $\alpha$ & noise scale $\sigma$ & Langevin step size $\epsilon$ & \#step \\ \midrule
Earthquake    &0.5   &0.2/0.25        & 0.005                          & 2000  \\
Crime         &0.5        &0.3/0.03  &0.005                          & 1000  \\
Football      &0.5 & 0.2/0.1       & 0.01& 2000\\ 
StackOverflow &0.2   &0.1        & 0.005                          & 2000  \\
Retweet       &0.5        &0.1  &0.0003                          & 2000  \\
Financial     &0.5 & 0.01      & 0.005                          & 2000  \\ 
MIMIC-II      &0.5   &0.005       & 0.002                          & 200  \\
\bottomrule
\end{tabular}%
\end{table}

\subsection{Computing Confidence Interval/Region}
Let's denote the $Q$ generated samples as $\{(t_i^j, \bm{x}_i^j, k_i^j)\}_{j=1}^Q$. For event time, which often follows long-tail distribution, we calculate the $q$-confidence level interval as $[0, t_i^q]$, with $t_i^q$ being the $q$-th quantile of sample times. For event location, we first obtain the estimated pdf through gaussian kernel density estimation. A threshold is then determined such that the region with a density exceeding this threshold satisfies the desired confidence level. This delineated region becomes the final confidence region.

\subsection{Baselines}
Most of the neural STPP baselines we consider do not inherently support marked STPP modeling. We augment them for marked STPP by parametrizing an intensity function for each mark based on their original approach.
We utilize multiple sampling methods to adapt to different baselines.
For all baselines except DSTPP, we sample event time by Langevin Dynamics from the learned intensities as in our method.
We sample event location for NJSDE by directly sampling from the learned gaussian distribution.
For NSTPP, we sample through their CNF procedure, that is, sampling starts from the initial distribution followed by the ODE progression.
Both NSMPP and DeepSTPP, which model joint intensity, utilize Langevin Dynamics for event location sampling.
DSTPP sampling adheres to its native diffusion sampling method.

\begin{table}[h]
\centering
\caption{Datasets statistics containing the name of the dataset, the number of event types, the number of events, and the average length per sequence}
\label{dataset}
\begin{tabular}{cccc}
\toprule
Dataset       & \#Type & \#Event  & Average Length \\ \midrule
Earthquake & 3    & 88064& 73\\
Crime       & 4& 35381& 141\\
Football     & 7& 67408& 688\\
StackOverflow & 22    & 480413  & 64             \\
Retweet       & 3     & 2173533 & 109            \\
Financial     & 2     & 414800  & 2074           \\
MIMIC-II      & 75    & 2419    & 4              \\ \bottomrule
\end{tabular}
\vspace{-2ex}
\end{table}

\section{Additional Results}
\subsection{Sample Visualization}
\begin{figure}[htb!]
\centering
\subfigure[Crime]{\label{sample_dis_crime_t}
\includegraphics[width=0.24\linewidth]{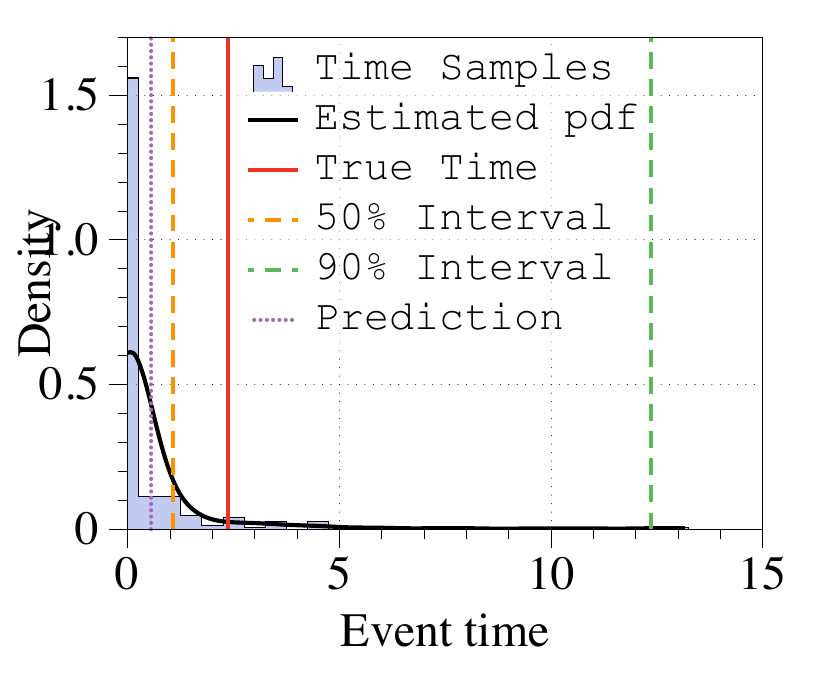}%
\includegraphics[width=0.24\linewidth]{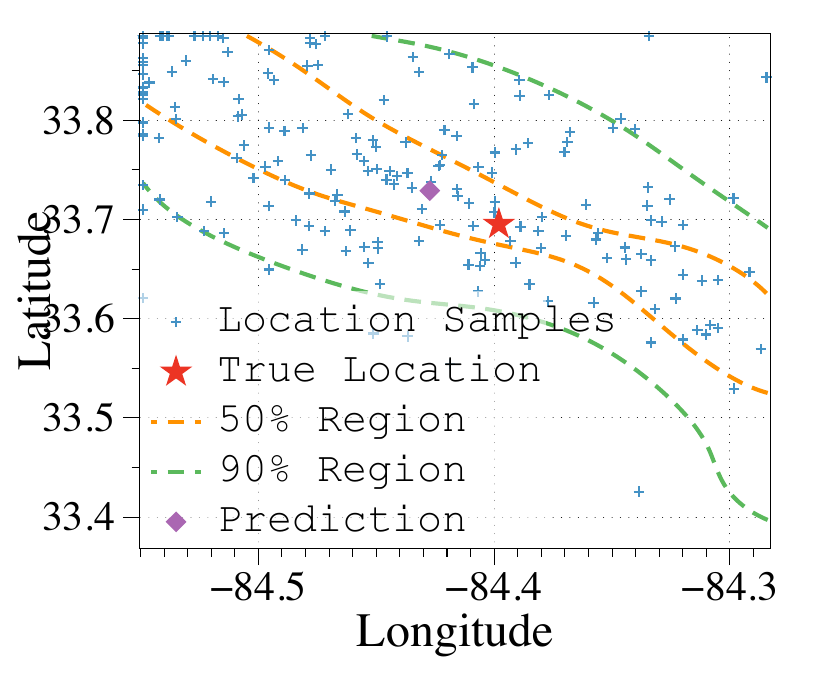}%
}
\subfigure[Football]{\label{sample_dis_football_t}
\includegraphics[width=0.24\linewidth]{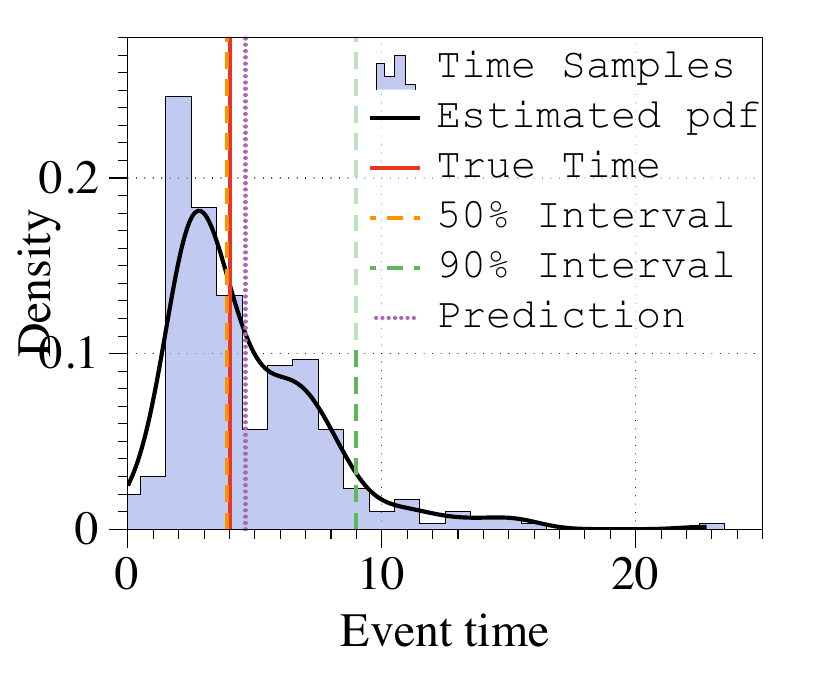}%
\includegraphics[width=0.24\linewidth]{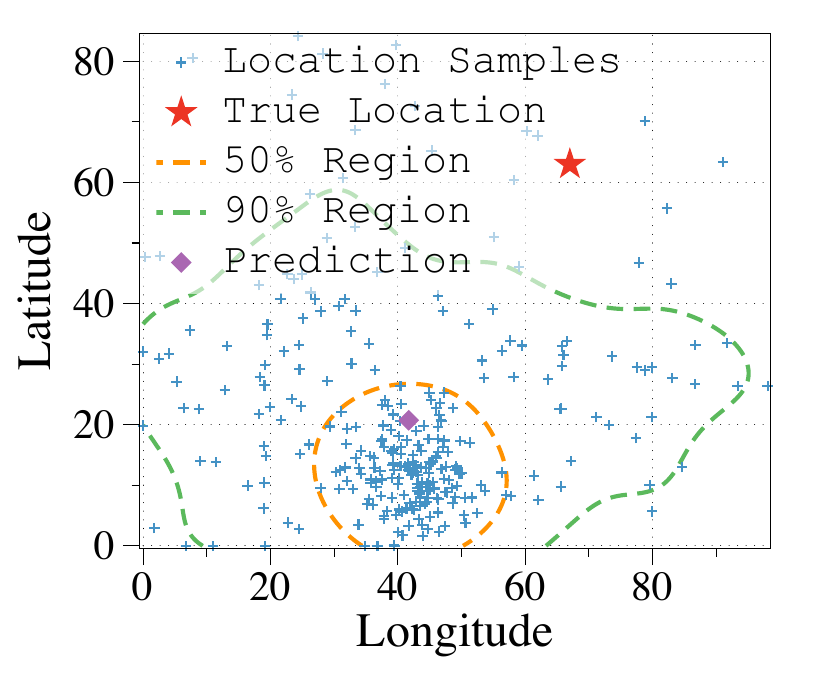}%
}
\vspace{-0.10in}
\caption{Distribution of samples' times and locations generated by SMASH and the ground truth on the Crime and Football dataset.}
\label{sample_dis2}
\end{figure}

We visualize the sample distribution for two randomly selected events from the Crime and Football datasets in Figure \ref{sample_dis2}. We can observe a long-tail distribution of event time and a wide-spread distribution of event location.

\subsection{Different Confidence Levels}
\begin{figure}[htb!]
\centering
\subfigure[Earthquake]{\label{allq_earth}
\includegraphics[width=0.24\linewidth]{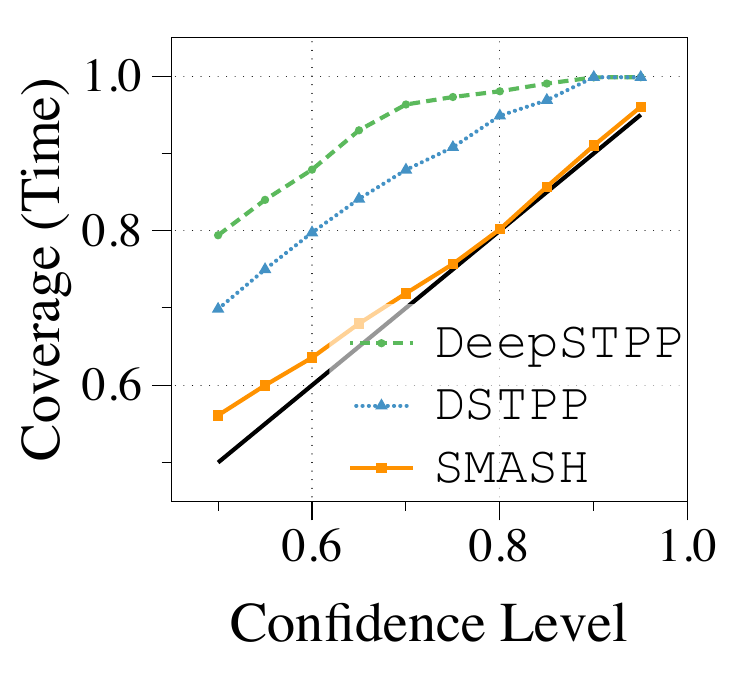}%
\includegraphics[width=0.24\linewidth]{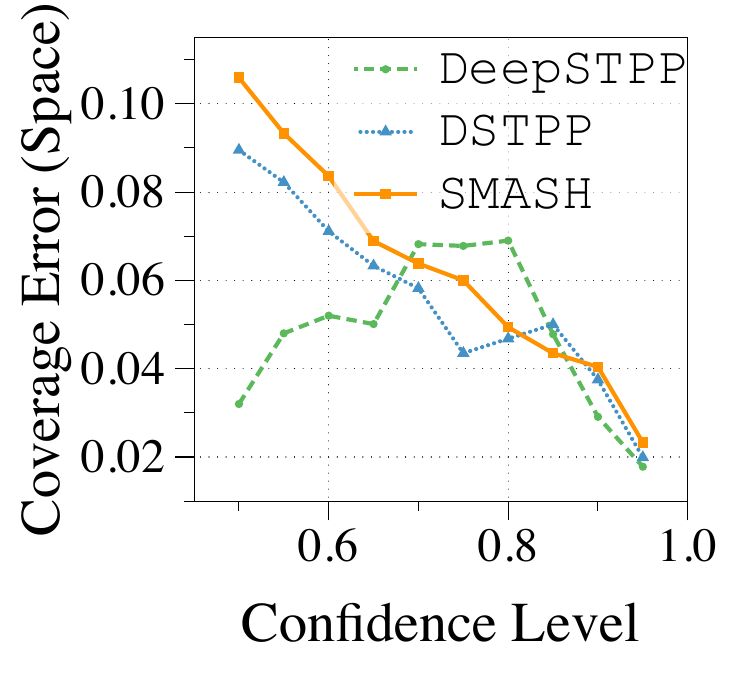}%
}
\subfigure[Football]{\label{allq_football}
\includegraphics[width=0.24\linewidth]{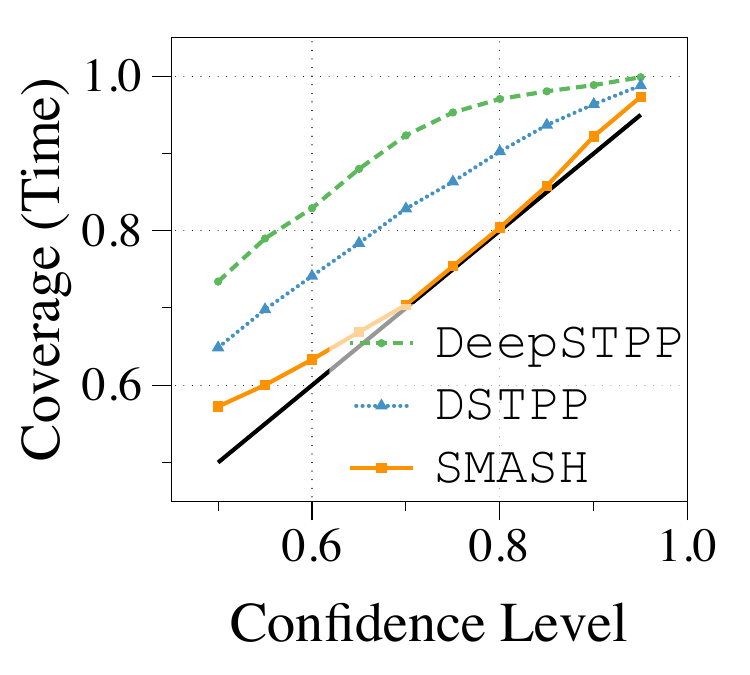}%
\includegraphics[width=0.24\linewidth]{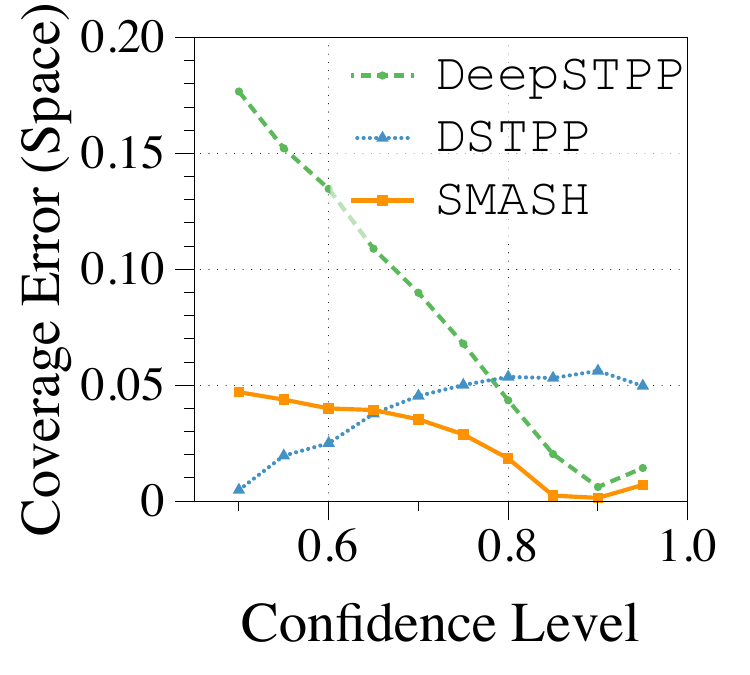}%
}
\vspace{-0.10in}
\caption{Comparison of coverage of different confidence levels on the Earthquake and Football dataset.}
\label{all_q_2}
\end{figure}

We display coverage and coverage error for different confidence levels on the Earthquake and Football datasets in Figure \ref{all_q_2}. For event time, SMASH consistently outperforms DeepSTPP and DSTPP across all confidence levels. For event location, SMASH achieves comparable coverage error with DSTPP on the Earthquake dataset and slightly better performance on the Football dataset.

%% file: main_arxiv.bbl
\begin{thebibliography}{}

\bibitem[Chen et~al., 2021]{ODESTPP_2021}
Chen, R. T.~Q., Amos, B., and Nickel, M. (2021).
\newblock Neural spatio-temporal point processes.
\newblock In {\em {ICLR}}. OpenReview.net.

\bibitem[Diggle, 2006]{STPP_review2006}
Diggle, P.~J. (2006).
\newblock Spatio-temporal point processes: methods and applications.
\newblock {\em Monographs on Statistics and Applied Probability}, 107:1.

\bibitem[Dong et~al., 2023]{DNSTPP_2023}
Dong, Z., Cheng, X., and Xie, Y. (2023).
\newblock Spatio-temporal point processes with deep non-stationary kernels.
\newblock In {\em {ICLR}}. OpenReview.net.

\bibitem[Du et~al., 2016]{rmtpp2016}
Du, N., Dai, H., Trivedi, R., Upadhyay, U., Gomez{-}Rodriguez, M., and Song, L.
  (2016).
\newblock Recurrent marked temporal point processes: Embedding event history to
  vector.
\newblock In {\em {KDD}}, pages 1555--1564. {ACM}.

\bibitem[Efron, 2011]{tweedie}
Efron, B. (2011).
\newblock Tweedie’s formula and selection bias.
\newblock {\em Journal of the American Statistical Association},
  106(496):1602--1614.
\newblock PMID: 22505788.

\bibitem[Gonz{\'a}lez et~al., 2016]{STPP_review2016}
Gonz{\'a}lez, J.~A., Rodr{\'\i}guez-Cort{\'e}s, F.~J., Cronie, O., and Mateu,
  J. (2016).
\newblock Spatio-temporal point process statistics: a review.
\newblock {\em Spatial Statistics}, 18:505--544.

\bibitem[Guo et~al., 2017]{calibration_ece}
Guo, C., Pleiss, G., Sun, Y., and Weinberger, K.~Q. (2017).
\newblock On calibration of modern neural networks.
\newblock In {\em {ICML}}, volume~70 of {\em Proceedings of Machine Learning
  Research}, pages 1321--1330. {PMLR}.

\bibitem[Guo et~al., 2018]{initiator2018}
Guo, R., Li, J., and Liu, H. (2018).
\newblock {INITIATOR:} noise-contrastive estimation for marked temporal point
  process.
\newblock In {\em {IJCAI}}, pages 2191--2197. ijcai.org.

\bibitem[Hawkes, 1971]{hawkes1971spectra}
Hawkes, A.~G. (1971).
\newblock Spectra of some self-exciting and mutually exciting point processes.
\newblock {\em Biometrika}, 58(1):83--90.

\bibitem[Hsieh et~al., 2018]{mirror}
Hsieh, Y., Kavis, A., Rolland, P., and Cevher, V. (2018).
\newblock Mirrored langevin dynamics.
\newblock In {\em NeurIPS}, pages 2883--2892.

\bibitem[Hyv{\"{a}}rinen, 2005]{score_matching2005}
Hyv{\"{a}}rinen, A. (2005).
\newblock Estimation of non-normalized statistical models by score matching.
\newblock {\em J. Mach. Learn. Res.}, 6:695--709.

\bibitem[Jia and Benson, 2019]{SDE_2019}
Jia, J. and Benson, A.~R. (2019).
\newblock Neural jump stochastic differential equations.
\newblock In {\em NeurIPS}, pages 9843--9854.

\bibitem[Kingman, 1992]{kingman1992poisson}
Kingman, J. F.~C. (1992).
\newblock {\em Poisson processes}, volume~3.
\newblock Clarendon Press.

\bibitem[Leskovec and Krevl, 2014]{so_data}
Leskovec, J. and Krevl, A. (2014).
\newblock {SNAP Datasets}: {Stanford} large network dataset collection.
\newblock \url{http://snap.stanford.edu/data}.

\bibitem[Li et~al., 2021]{STPP_covid2021}
Li, S., Wang, L., Chen, X., Fang, Y., and Song, Y. (2021).
\newblock Understanding the spread of {COVID-19} epidemic: {A} spatio-temporal
  point process view.
\newblock {\em CoRR}, abs/2106.13097.

\bibitem[Li et~al., 2018]{TPPRL2018}
Li, S., Xiao, S., Zhu, S., Du, N., Xie, Y., and Song, L. (2018).
\newblock Learning temporal point processes via reinforcement learning.
\newblock In {\em NeurIPS}, pages 10804--10814.

\bibitem[Li et~al., 2023]{smurf2023}
Li, Z., Xu, Y., Zuo, S., Jiang, H., Zhang, C., Zhao, T., and Zha, H. (2023).
\newblock {SMURF-THP:} score matching-based uncertainty quantification for
  transformer hawkes process.
\newblock In {\em {ICML}}, volume 202 of {\em Proceedings of Machine Learning
  Research}, pages 20210--20220. {PMLR}.

\bibitem[Mei and Eisner, 2017]{nhp2017}
Mei, H. and Eisner, J. (2017).
\newblock The neural hawkes process: {A} neurally self-modulating multivariate
  point process.
\newblock In {\em {NIPS}}, pages 6754--6764.

\bibitem[Mei et~al., 2020]{ncetpp2020}
Mei, H., Wan, T., and Eisner, J. (2020).
\newblock Noise-contrastive estimation for multivariate point processes.
\newblock In {\em NeurIPS}.

\bibitem[Meng et~al., 2022]{meng2022concrete}
Meng, C., Choi, K., Song, J., and Ermon, S. (2022).
\newblock Concrete score matching: Generalized score matching for discrete
  data.
\newblock {\em arXiv preprint arXiv:2211.00802}.

\bibitem[Song and Ermon, 2019]{SMLD}
Song, Y. and Ermon, S. (2019).
\newblock Generative modeling by estimating gradients of the data distribution.
\newblock In {\em NeurIPS}, pages 11895--11907.

\bibitem[Sullivan, 2015]{UQ_book}
Sullivan, T.~J. (2015).
\newblock {\em Introduction to uncertainty quantification}, volume~63.
\newblock Springer.

\bibitem[Tagliazucchi et~al., 2012]{STPP_brain2012}
Tagliazucchi, E., Balenzuela, P., Fraiman, D., and Chialvo, D.~R. (2012).
\newblock Criticality in large-scale brain fmri dynamics unveiled by a novel
  point process analysis.
\newblock {\em Frontiers in physiology}, 3:15.

\bibitem[Upadhyay et~al., 2018]{RLPP2018}
Upadhyay, U., De, A., and Rodriguez, M.~G. (2018).
\newblock Deep reinforcement learning of marked temporal point processes.
\newblock In {\em NeurIPS}, pages 3172--3182.

\bibitem[Vaswani et~al., 2017]{attention}
Vaswani, A., Shazeer, N., Parmar, N., Uszkoreit, J., Jones, L., Gomez, A.~N.,
  Kaiser, L., and Polosukhin, I. (2017).
\newblock Attention is all you need.
\newblock In {\em {NIPS}}, pages 5998--6008.

\bibitem[Vincent, 2011]{denoise2011}
Vincent, P. (2011).
\newblock A connection between score matching and denoising autoencoders.
\newblock {\em Neural Comput.}, 23(7):1661--1674.

\bibitem[Xiao et~al., 2017]{DIS2017}
Xiao, S., Yan, J., Yang, X., Zha, H., and Chu, S.~M. (2017).
\newblock Modeling the intensity function of point process via recurrent neural
  networks.
\newblock In {\em {AAAI}}, pages 1597--1603. {AAAI} Press.

\bibitem[Yeung et~al., 2023]{football}
Yeung, C.~C., Sit, T., and Fujii, K. (2023).
\newblock Transformer-based neural marked spatio temporal point process model
  for football match events analysis.
\newblock {\em arXiv preprint arXiv:2302.09276}.

\bibitem[Yu et~al., 2019]{gsm2019}
Yu, S., Drton, M., and Shojaie, A. (2019).
\newblock Generalized score matching for non-negative data.
\newblock {\em J. Mach. Learn. Res.}, 20:76:1--76:70.

\bibitem[Yu et~al., 2022]{gsm2022}
Yu, S., Drton, M., and Shojaie, A. (2022).
\newblock Generalized score matching for general domains.
\newblock {\em Information and Inference: A Journal of the IMA},
  11(2):739--780.

\bibitem[Yuan et~al., 2023]{DSTPP_2023}
Yuan, Y., Ding, J., Shao, C., Jin, D., and Li, Y. (2023).
\newblock Spatio-temporal diffusion point processes.
\newblock In {\em {KDD}}, pages 3173--3184. {ACM}.

\bibitem[Zhang et~al., 2020]{sahp2020}
Zhang, Q., Lipani, A., Kirnap, {\"{O}}., and Yilmaz, E. (2020).
\newblock Self-attentive hawkes process.
\newblock In {\em {ICML}}, volume 119 of {\em Proceedings of Machine Learning
  Research}, pages 11183--11193. {PMLR}.

\bibitem[Zhao et~al., 2015]{retweet_data}
Zhao, Q., Erdogdu, M.~A., He, H.~Y., Rajaraman, A., and Leskovec, J. (2015).
\newblock {SEISMIC:} {A} self-exciting point process model for predicting tweet
  popularity.
\newblock In {\em {KDD}}, pages 1513--1522. {ACM}.

\bibitem[Zhou et~al., 2022]{DeepSTPP_2022}
Zhou, Z., Yang, X., Rossi, R.~A., Zhao, H., and Yu, R. (2022).
\newblock Neural point process for learning spatiotemporal event dynamics.
\newblock In {\em {L4DC}}, volume 168 of {\em Proceedings of Machine Learning
  Research}, pages 777--789. {PMLR}.

\bibitem[Zhu et~al., 2022]{SpecSTPP_2022}
Zhu, S., Wang, H., Dong, Z., Cheng, X., and Xie, Y. (2022).
\newblock Neural spectral marked point processes.
\newblock In {\em {ICLR}}. OpenReview.net.

\bibitem[Zuo et~al., 2020]{thp2020}
Zuo, S., Jiang, H., Li, Z., Zhao, T., and Zha, H. (2020).
\newblock Transformer hawkes process.
\newblock In {\em {ICML}}, volume 119 of {\em Proceedings of Machine Learning
  Research}, pages 11692--11702. {PMLR}.

\end{thebibliography}
